\documentclass[11pt, letterpaper, logo, onecolumn, copyright]{main}

\usepackage[authoryear, sort&compress, round]{natbib}

\usepackage[most, breakable, skins]{tcolorbox}

\tcbuselibrary{skins}
\usepackage{lipsum}
\usepackage{tabularx}
\usepackage{afterpage}
\usepackage{booktabs}
\usepackage{subcaption}
\usepackage{makecell}
\usepackage{multirow}
\usepackage{bm}
\usepackage{multicol}
\usepackage{array}
\usepackage{float}
\usepackage{listings, listings-rust}
\usepackage{fontawesome5}
\usepackage{hyperref}
\usepackage{amssymb,graphicx}
\usepackage[dvipsnames]{xcolor}
\usepackage{cleveref}
\usepackage{longtable}
\usepackage{pdflscape}
\usepackage{adjustbox}
\usepackage{nicematrix}
\usepackage{CJKutf8}
\usepackage{ragged2e}
\usepackage{colortbl}
\usepackage{enumitem}
\usepackage[ruled,linesnumbered]{algorithm2e}
\usepackage{bbm}
\usepackage{pifont}
\usepackage[htt]{hyphenat}
\usepackage{amsmath}
\usepackage{amsthm}  
\usepackage{mathrsfs} 
\usepackage{circledsteps}
\usepackage{diagbox}
\usepackage{bookmark}

\usepackage{wrapfig}
\usepackage{xspace}
\usepackage{tikz}
\usepackage[normalem]{ulem}

\usepackage{pgfplots}
\usepackage{pgfplotstable}
\pgfplotsset{compat=1.18}

\usepackage[most]{tcolorbox}
\definecolor{lightblue}{RGB}{210, 220, 250}

\definecolor{medgray55}{gray}{0.55}
\definecolor{medgray}{gray}{0.7}
\definecolor{litegray}{gray}{0.9}
\definecolor{gblue}{RGB}{210, 227, 252}
\definecolor{gred}{RGB}{250, 210, 207}
\definecolor{gyellow}{RGB}{254, 239, 195}
\definecolor{ggreen}{RGB}{206, 234, 214}
\definecolor{gorange}{RGB}{254, 223, 200}

\definecolor{gblue9}{RGB}{23, 78, 166}
\definecolor{gred9}{RGB}{165, 14, 14}
\definecolor{gyellow9}{RGB}{227, 116, 0}
\definecolor{ggreen9}{RGB}{13, 101, 45}
\definecolor{gorange9}{RGB}{176, 96, 0}

\definecolor{myblue}{rgb}{0,0,1}
\definecolor{myred}{rgb}{1,0,0}
\definecolor{mylightgray}{gray}{0.95}
\definecolor{myCite}{HTML}{1C4587}

\definecolor{highlightblue}{HTML}{185ABC}
\definecolor{cellHighlight}{HTML}{dbefff}
\definecolor{lightgray}{RGB}{211, 211, 211}
\definecolor{lightfont}{gray}{0.3}

\usepackage{minitoc}

\noptcrule


\newcolumntype{L}[1]{>{\raggedright\let\newline\\\arraybackslash\hspace{0pt}}m{#1}}
\newcolumntype{C}[1]{>{\centering}m{#1}}

\newcolumntype{R}[1]{>{\raggedleft\let\newline\\\arraybackslash\hspace{0pt}}m{#1}}

\definecolor{ao}{rgb}{0.0, 0.0, 1.0}

\newcommand\vcent[1]{\vcenter{\hbox{#1}}}

\newcommand\loudspeaker[1][3]{\ensuremath{\vcent{\rule{.6ex}{.6ex}}\kern-.5ex
  \vcent{\scalebox{.6}[1]{\rotatebox[origin=center]{90}{$\blacktriangle$}}}
  \ifnum#1>0\relax\kern.05ex\vcent{\scalebox{.4}{\ttfamily)}}
  \ifnum#1>1\relax\kern-.4ex\vcent{\scalebox{.56}{\ttfamily)}}
  \ifnum#1>2\relax\kern-.55ex\vcent{\scalebox{.7}{\ttfamily)}}
  \fi\fi\fi}
}

\makeatletter
\renewcommand\subparagraph{
 \@startsection {subparagraph}{5}{\z@ }{3.25ex \@plus 1ex
 \@minus .2ex}{-1em}{\normalfont \normalsize \bfseries }}
\makeatother

\let\cite\citep
\hypersetup{
  citecolor = myCite,  
  linkcolor = myCite,   
  urlcolor  = myCite
}

\usepackage{algpseudocode}
\usepackage{arydshln}

\title{AgentPRM: Process Reward Models for LLM Agents via Step-Wise Promise and Progress}

\author{
Zhiheng Xi$^1$$^{* \dag}$, Chenyang Liao$^1$$^*$, Guanyu Li$^1$, Yajie Yang$^1$, Wenxiang Chen$^1$, \\
\textbf{Zhihao Zhang$^1$, Binghai Wang$^1$, Senjie Jin$^1$, Yuhao Zhou$^1$, Jian Guan$^2$,} \\
\textbf{Wei Wu$^2$, Tao Ji$^1$, Tao Gui$^1$$^{\dag}$, Qi Zhang$^1$$^{\dag}$,  Xuanjing Huang$^1$$^{\dag}$}\\

$^1$Fudan University $^2$Ant Group  \\
\texttt{zhxi22@m.fudan.edu.cn, \{tgui,xjhuang\}@fudan.edu.cn} 
}
\vspace{-50pt}
\begin{abstract}
Despite rapid development, large language models (LLMs) still encounter challenges in multi-turn decision-making tasks (i.e., agent tasks) like web shopping and browser navigation, which require making a sequence of intelligent decisions based on environmental feedback. Previous work for LLM agents typically relies on elaborate prompt engineering or fine-tuning with expert trajectories to improve performance. In this work, we take a different perspective: we explore constructing process reward models (PRMs) to evaluate each decision and guide the agent's decision-making process. Unlike LLM reasoning, where each step is scored based on correctness, actions in agent tasks do not have a clear-cut correctness. Instead, they should be evaluated based on their proximity to the goal and the progress they have made. Building on this insight, we propose a re-defined PRM for agent tasks, named AgentPRM, to capture both the interdependence between sequential decisions and their contribution to the final goal. This enables better progress tracking and exploration-exploitation balance. To scalably obtain labeled data for training AgentPRM, we employ a Temporal Difference-based (TD-based) estimation method combined with Generalized Advantage Estimation (GAE), which proves more sample-efficient than prior methods. Extensive experiments across different agentic tasks show that AgentPRM is over $8\times$ more compute-efficient than baselines, and it demonstrates robust improvement when scaling up test-time compute. Moreover, we perform detailed analyses to show how our method works and offer more insights, e.g., applying AgentPRM to the reinforcement learning of LLM agents.
\end{abstract}

\begin{document}

\doparttoc
\faketableofcontents

\begingroup
  \renewcommand\thefootnote{}
  \footnote{\textsuperscript{*}Equal contribution.
            \textsuperscript{\dag}Corresponding authors.}
  \addtocounter{footnote}{-1}
\endgroup

\vspace{-30pt}
\maketitle

\vspace{-15pt}
\section{Introduction}

The advent of large language models (LLMs) has resulted in significant advances in a variety of natural language processing tasks, including text generation \citep{keskar2019ctrlconditionaltransformerlanguage, dathathri2020plugplaylanguagemodels, 10.1145/3617680, 10.1145/3649449}, summarization \citep{tang2023evaluating, 10.1162/tacla00632, van2024adapted, zhang2025comprehensivesurveyprocessorientedautomatic}, translation \citep{moslem2023adaptivemachinetranslationlarge, wang2023documentlevelmachinetranslationlarge, pmlr-v202-zhang23m, xu2024paradigmshiftmachinetranslation, 10.1162/tacla00642}, and reasoning \citep{qwen2.5, openai2024GPT4o, DBLP:journals/corr/abs-2312-11805}. Despite these advancements, LLMs still encounter considerable challenges in multi-turn decision-making tasks (i.e., agent tasks) such as web shopping \citep{DBLP:conf/nips/Yao0YN22, zhang2025shopr1rewardingllmssimulate}, browser navigation \citep{DBLP:conf/iclr/ZhouX0ZLSCOBF0N24, deng2023mind2webgeneralistagentweb, xu2021groundingopendomaininstructionsautomate, koh2024visualwebarenaevaluatingmultimodalagents}, and digital games \citep{DBLP:conf/iclr/Chevalier-Boisvert19, DBLP:conf/naacl/PrasadKHCSBK24, wang2025digitalplayerevaluatinglarge, park2025orakfoundationalbenchmarktraining, hu2025gamearenaevaluatingllmreasoning}, where models must make a series of intelligent decisions based on feedback from the environment \citep{DBLP:journals/corr/abs-2309-07864, DBLP:conf/acl/ZengLLWLD024}. These models are referred to as LLM agents \citep{DBLP:journals/corr/abs-2309-07864}.

\vspace{10pt}

\begin{wrapfigure}[23]{r}[0pt]{0pt}
    \centering
    \includegraphics[width=0.55\linewidth]{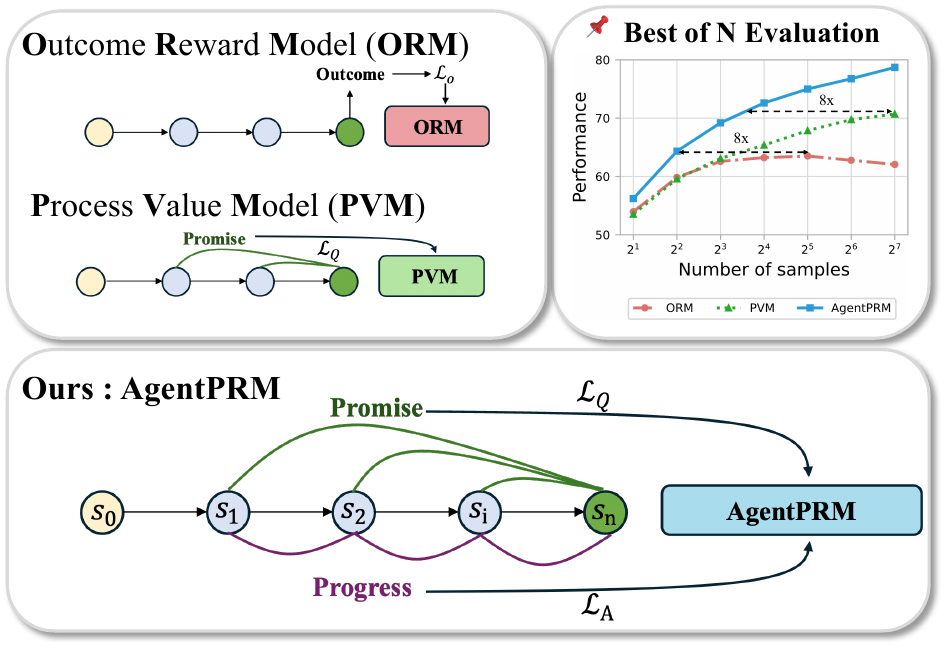}
    \caption{
    Comparison of AgentPRM and baselines, and the Best-of-N results. \underline{Upper Left}: Baseline reward models. ORMs focus on the final outcome reward; PVMs focus on promise of each step only. \underline{Bottom}: Our AgentPRM that captures both the promise and progress of each step. \underline{Upper Right}: Average Best-of-N performance of three agent tasks. AgentPRM outperforms other baselines, and it demonstrates a more stable and robust improvement trend as inference compute scaling.
    }
    \label{fig: vice}
\end{wrapfigure}

The agent tasks are inherently dynamic and context-sensitive, setting them apart from traditional static tasks \citep{DBLP:conf/iclr/0036YZXLL0DMYZ024, DBLP:conf/acl/ChenLWZLLCZ24, DBLP:journals/corr/abs-2406-04151}. Achieving effective performance in these tasks requires models not only to comprehend task-related knowledge and interpret environmental cues, but also to engage in forward-looking planning to anticipate future consequences of their decisions \citep{DBLP:journals/corr/abs-2309-07864, DBLP:journals/fcsc/WangMFZYZCTCLZWW24}.

Prior work has sought to enhance LLMs for agent tasks using approaches such as supervised fine-tuning \citep{DBLP:conf/acl/ZengLLWLD024, DBLP:conf/acl/ChenLWZLLCZ24} and prompt engineering \citep{DBLP:conf/iclr/YaoZYDSN023, DBLP:journals/corr/abs-2308-05960, DBLP:conf/nips/ShinnCGNY23}. 
Supervised fine-tuning methods rely on expert-labeled data, which are scarce and hard to collect, limiting scalability. Prompt engineering typically leverages commercial models like GPT-4o \citep{openai2024GPT4o} to achieve satisfactory performance, but is hindered by API constraints, making it both costly and inflexible for customization \citep{DBLP:journals/corr/abs-2306-02224, DBLP:journals/corr/abs-2407-01476}. Another promising direction involves self-improvement that trains models by leveraging successful trajectories they explored \citep{DBLP:journals/corr/abs-2406-04151, aksitov2023restmeetsreactselfimprovement, song2024trialerrorexplorationbasedtrajectory, yang2024reactmeetsactrelanguage, tao2024surveyselfevolutionlargelanguage}. However, it relies on outcome-based feedback, which does not offer sufficient insight into the value of individual decisions made by the model and, in turn, leads to a performance bottleneck \citep{ding2025mitigatingtailnarrowingllm}.

To this end, we draw inspiration from process supervision in LLM reasoning and explore the use of process reward models (PRMs) to guide the search and exploration of LLM agents \citep{DBLP:journals/corr/abs-2211-14275,DBLP:conf/acl/WangLSXDLCWS24,DBLP:conf/iclr/LightmanKBEBLLS24,DBLP:journals/corr/abs-2410-08146,DBLP:journals/corr/abs-2410-11287}. While PRMs have proven effective in reasoning tasks to evaluate individual steps and guide the decoding process of LLMs, they face unique challenges in agent tasks:

\begin{enumerate}
    \item Different from LLM reasoning, actions in agent tasks lack a clear-cut "correctness" metric, making the evaluation non-trivial \citep{DBLP:conf/nips/Yao0YN22,DBLP:conf/iclr/ZhouX0ZLSCOBF0N24}. 
    \item Existing process supervision methods treat each step independently, overlooking the sequential dependencies between decisions within a trajectory, which is inconsistent with the inherently sequential nature of agent tasks \citep{DBLP:journals/corr/abs-2410-11287}.
    \item Previous methods for training PRMs often depend on either expert annotations or extensive Monte Carlo-based (MC-based) sampling for estimation, both of which are costly in real-world scenarios \citep{DBLP:conf/acl/WangLSXDLCWS24,DBLP:journals/corr/abs-2406-06592}. 
\end{enumerate}

\begin{figure*}[t]
    \centering
    \includegraphics[width=0.99\linewidth]{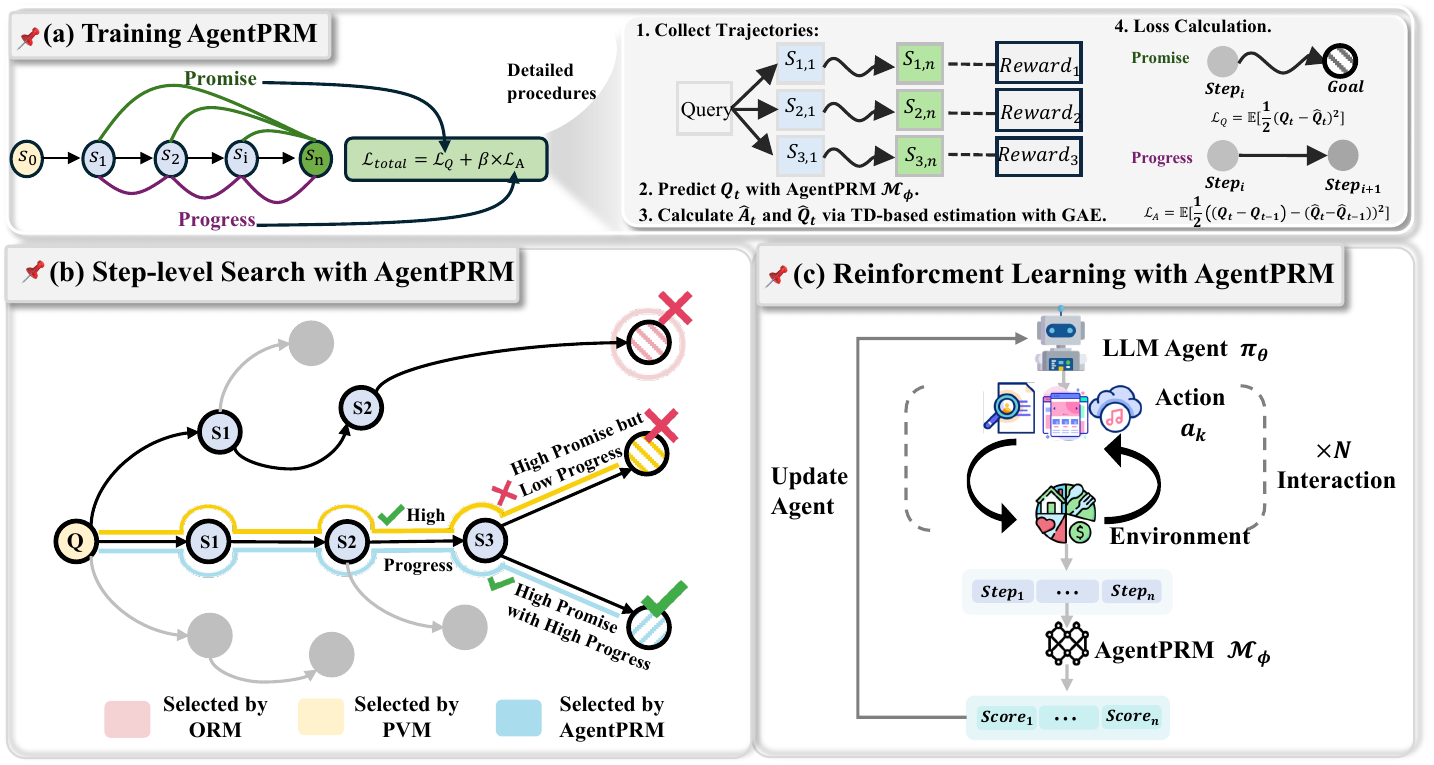}
    \caption{
    Overview of the training and the application of AgentPRM. (a): The training objective and the detailed training procedures of AgentPRM. We take into account both the promise (probability of each step achieving the goal) and the progress (the interdependence between sequential steps). (b): With AgentPRM, we perform step-level beam search to guide the LLM agent toward the goal. (c): AgentPRM can be integrated into the reinforcement learning process of LLM agents seamlessly.
    }
    \label{fig: AgentPRM main fig}
\end{figure*}

In this work, we propose AgentPRM to address the challenges (Figure \ref{fig: vice} and Figure \ref{fig: AgentPRM main fig}). Our core insight is to \emph{evaluate the proximity of each step to the goal state and tracks the progress made by LLMs}. Specifically, AgentPRM predicts the contribution of each decision to the final goal and captures the interdependencies between sequential decisions, thereby enabling more effective progress tracking and optimizing the balance between exploration and exploitation. 
To scale the training data acquisition efficiently, we employ an automated Temporal Difference-based (TD-based) method with Generalized Advantage Estimation (GAE) \citep{DBLP:journals/corr/SchulmanMLJA15}, which is more efficient than previous MC-based methods \citep{DBLP:conf/acl/WangLSXDLCWS24}, and provides a better trade-off between variance and bias in estimation \citep{DBLP:journals/corr/SchulmanMLJA15}. 

Extensive experiments across various models and tasks show that AgentPRM consistently outperforms baselines in both performance and compute efficiency. For instance, with Qwen2.5-3B, AgentPRM achieves over $8\times$ greater compute efficiency compared to baselines across three agent tasks and multiple sampling strategies. Additionally, it demonstrates a more stable and robust improvement trend as inference compute scales. Further analyses, including its application to reinforcement learning (\S \ref{sec:Applying AgentPRM to Reinforcement Learning}) and a comparison of sampling efficiency with baselines (\S \ref{sec: Comparing Sampling Efficiency of Our Method with MC-based Estimation}), are also provided to offer more insights.

To summarize, this work makes the following key contributions:
\begin{itemize}
    \item Drawing inspiration from the nature of agentic tasks, we propose AgentPRM, a novel process reward model for LLM agents that simultaneously captures both the immediate progress and the long-term promise of each decision.
    \item We propose an automated, scalable method, i.e., TD-based estimation with GAE, for training AgentPRM, which is much more efficient than traditional MC-based methods.
    \item Through extensive experiments across diverse agentic tasks, we demonstrate that AgentPRM achieves over $8\times$ more compute-efficient than baselines, and it demonstrates robust improvement when scaling up test-time compute. We also perform in-depth ablation and analyses to show how it works and provide more insights.
\end{itemize}

\section{Preliminary and Background}\label{sec: Preliminary and Background}

An agent task can be formalized as a Partially Observable Markov Decision Process (POMDP) \\ $(\mathcal{U, S, A, O, T}, r)$ \citep{DBLP:journals/corr/abs-2406-04151, DBLP:conf/aaaifs/HausknechtS15}, where $\mathcal{U}$ is the instruction space, $\mathcal{S}$ is the state space, $\mathcal{A}$ is the action space, $\mathcal{O}$ is the observation space, $\mathcal{T: S \times A \to S}$ is the deterministic state transition function, and $r: \mathcal{S \times A} \to \mathbb{R}$ is the reward function. Given a task instruction $u \in \mathcal{U}$, the initial observation $o_0 \in \mathcal{O}$, and the initial state $s_0 = \{u, o_0\}$, the agent selects an action $a_0 \sim \pi_\theta(\cdot | s_0) \in \mathcal{A}$ under a policy $\pi_\theta$ parameterized by $\theta$. The environment then returns an observation $o_1 \in O$, yielding the next state $s_1 = \{u, s_0, a_0, o_1\}$ via $\mathcal{T}$. Following the process, the agent proposes a sequence of actions $\{a_t\}_{t=0}^T$, where $T$ is the number of steps, to interact with the environment until the task is completed or the maximum number of steps is reached: $\tau = (u, o_0, a_0, o_1, \cdots, o_{T}, a_T)$. Then for a language model, the agent task can be formalized as:
\begin{equation}
    \pi_\theta(\tau | s_0) = \prod_{t=0}^T \pi_\theta(a_t | s_t),
\end{equation}
where $s_t$ represents the interaction history up to timestep $t$. Finally, the environment $e$ provides an outcome reward $r(u, \tau) \in [0, 1]$ to describe the completion of the agent task.

\paragraph{Outcome Reward Model (ORM)}
An ORM $ r_\textnormal{orm} $ takes a trajectory \( \tau \) as input and predicts whether it satisfies the task instruction \( u \) \citep{DBLP:journals/corr/abs-2211-14275,DBLP:conf/nips/Ouyang0JAWMZASR22}. ORMs are trained on data sampled from \( \pi_\theta \), where each instruction–trajectory pair is labeled by the corresponding outcome reward \( r(u, \tau) \) \citep{DBLP:conf/nips/Ouyang0JAWMZASR22, DBLP:journals/corr/abs-2110-14168, DBLP:journals/corr/abs-2410-18451}.

\paragraph{Process Reward Model (PRM)}
A PRM evaluates actions or intermediate states along a trajectory \citep{DBLP:conf/acl/WangLSXDLCWS24, DBLP:conf/iclr/LightmanKBEBLLS24, DBLP:journals/corr/abs-2406-03816}. In the field of LLM reasoning, the scoring criterion typically involves the correctness of individual steps. However, this is not suitable for the agent tasks we are studying, which will be elaborated in \S \ref{sec: Methodology} and we will provide appropriate evaluation criteria. PRMs are trained using step-level annotations that assign labels to intermediate actions, and models are optimized to predict these labels \citep{DBLP:conf/acl/WangLSXDLCWS24, DBLP:conf/iclr/LightmanKBEBLLS24, DBLP:journals/corr/abs-2406-06592}.

\paragraph{Best-of-N (BoN) with reward models}
Given a larger inference budget, Best-of-N (BoN) can be applied to improve performance \citep{DBLP:journals/corr/abs-2307-09288}. Specifically, the policy \( \pi_\theta \) samples \( N \) trajectories \( \{\tau_i\}_{i=1}^N \), which are then evaluated by a reward model. The highest-scoring trajectory is selected as the final output. Note that BoN can also be executed with PRMs \citep{DBLP:conf/iclr/LightmanKBEBLLS24}. In our setting, the score of the final step is used to represent trajectory quality.

\paragraph{Search with process reward models}
During the inference phase, we can conduct step-level search against PRMs for agent tasks (Figure \ref{fig: AgentPRM main fig}(b)). Among the various step-level search algorithms, beam search is a widely used due to its balance between performance and efficiency \citep{DBLP:journals/corr/abs-2406-03816, DBLP:conf/nips/ChenL0024a}. In each iteration, beam search expands $M$ candidate actions per node, scores them with a PRM, and retains the top $N$ candidates. The trajectory ending in the highest-scoring terminal state is returned as the final solution. The algorithm of beam search is summarized in Algorithm \ref{algo: Beam search with AgentPRM} of Appendix \ref{app: Algorithm}.

\section{Methodology}\label{sec: Methodology}

\subsection{Motivation}

In LLM reasoning, researchers train PRMs by collecting annotated data to score each step based on its correctness. However, for agent tasks, three key challenges arise:

\begin{enumerate}
    \item Decisions in agent tasks do not have a clear-cut correctness, making evaluation non-trivial. For example, in web navigation, if the model makes a poor decision by clicking a button and navigating to a new page, it can immediately correct this by using the back button to return to the previous state \citep{DBLP:conf/nips/Yao0YN22, DBLP:conf/iclr/ZhouX0ZLSCOBF0N24}.
    \item Previous PRMs typically treat each state independently, without considering the dependencies between consecutive decisions \citep{DBLP:journals/corr/abs-2410-08146, DBLP:journals/corr/abs-2410-11287}. However, in agent tasks, the decisions at each step are interconnected, forming a chain of dependencies, where each decision influences subsequent decisions and ultimately the outcome \citep{DBLP:journals/corr/abs-2309-07864, DBLP:journals/corr/abs-2406-04151, DBLP:conf/nips/Yao0YN22, DBLP:conf/iclr/Chevalier-Boisvert19}.
    \item Previous methods for training PRMs often require expert annotations or a large amount of Monte Carlo-based (MC-based) sampling for estimation, both of which are costly \citep{DBLP:conf/acl/WangLSXDLCWS24, DBLP:conf/iclr/LightmanKBEBLLS24, DBLP:journals/corr/abs-2406-06592}. Moreover, MC-basd estimation may lead to high variance \citep{sutton2018reinforcement}.
\end{enumerate}

Given these challenges, our work focuses on two critical research questions: \emph{\textbf{RQ1: how to define appropriate rewards for decisions}} and \emph{\textbf{RQ2: how to efficiently and reliably train process reward models}}.

\subsection{AgentPRM: Re-Defining Process Rewards for LLM Agents}
To answer \textbf{\emph{RQ1}}, we argue that \emph{a good process reward model for agent tasks must consider both the probability that each step advances toward the goal (promise) and the interdependence among sequential steps (progress)}. Based on this insight, we propose the re-defined PRM for LLM agents in this section.

\subsubsection{Measuring expected future success probability with value functions.}
An agent task typically requires making a sequence of intelligent decisions aimed at reaching the goal state. Conceptually, this requires evaluating whether a decision brings the state closer to the goal \citep{DBLP:journals/corr/abs-2309-07864, DBLP:conf/iclr/0036YZXLL0DMYZ024, DBLP:conf/nips/Yao0YN22}. In RL, this is often defined as the action-value function \( Q^{\pi}(s_t,a_t) \) \citep{sutton2018reinforcement}, which measures the expected future success probability after taking a particular action $a_t$ based on state $s_t$:
\begin{equation}
    Q^{\pi}(s_t,a_t)= \mathbb{E}_{\tau \sim \pi(\cdot|s_t,a_t)} \left[ r(u,\tau) \right].
\end{equation}
Similarly, we can define the state-value function \( V(s_t) \) \citep{sutton2018reinforcement} with:
\begin{equation}
    V^{\pi}(s_t) = \mathbb{E}_{a_t \sim \pi(\cdot|s_t)} \left[ Q^{\pi}(s_t,a_t) \right].
\end{equation}
Now, given annotated labels for each state-action pair, \(\mathcal{D}_{Q} = \{s_t,a_t, \allowbreak \hat{Q}(s_t,a_t)\}\), we train our PRM $\mathcal{M}_\phi$ parameterized by $\phi$ to predict the action value with mean squared error (MSE) loss \citep{DBLP:conf/nips/ChenL0024a}:
\begin{equation}
    \mathcal{L}_Q(\phi) = \mathbb{E}_{s_t,a_t \sim \mathcal{D}_{Q}} \left[ \frac{1}{2}(\mathcal{M}_\phi(s_t,a_t) - \hat{Q}(s_t,a_t))^2 \right].
\end{equation}
After training, Based on the predictions of $\mathcal{M}_\phi$, we can perform inference-time search or BoN. 

\subsubsection{Capturing dependencies between steps with advantages.}
Nevertheless, the aforementioned $\mathcal{M}_\phi$ only considers the actions' contribution to the final goal, and fails to effectively capture the relationships and dependencies between consecutive states or decisions \citep{DBLP:journals/corr/abs-2410-08146, DBLP:journals/corr/abs-2410-11287}. In other words, it primarily measures promise but not progress. This often leads to excessive exploitation without a sufficient balance of exploration \citep{DBLP:journals/corr/abs-2410-08146,DBLP:journals/corr/abs-2408-03314}. However, in many agent tasks, models need sufficient exploration to successfully achieve the final goal \citep{DBLP:conf/nips/Yao0YN22, DBLP:conf/iclr/ZhouX0ZLSCOBF0N24, DBLP:conf/iclr/Chevalier-Boisvert19}. For example, in a web navigation task, the model needs to first navigate to the login page to log in, and then return to the current page to post a comment. Although the action of entering the login page may temporarily move the model away from the target page, it is still crucial because it is a necessary step to log in before posting.

Therefore, we argue that process rewards should not only measure the promise of success, but also capture the local progress between actions. This perspective aligns with RL \citep{sutton1999policy}, where advantage functions quantify the relative improvement in success likelihood resulting from an action:
\begin{equation}
    A^{\pi}(s_t,a_t) = Q^{\pi}(s_t,a_t) - V^{\pi}(s_t).
\end{equation}
The value of \( A^{\pi}(s_t,a_t) \) can be either positive or negative. A positive advantage indicates that the current action contributes to progress, whereas a negative advantage suggests the opposite.

Hence, to train our model $\mathcal{M}_\phi$ to account for both progress and dependencies between actions, we introduce a distinct loss term to fit the advantage:
\begin{equation}
    \begin{aligned}
    \mathcal{L}_A(\phi) = \mathbb{E}_{s_t,a_t \sim D_Q} \Big[(A_\phi(s_t, a_t) - \hat{A}(s_t,a_t))^2\Big],
    \end{aligned}
    \label{eqn: loss of advantage}
\end{equation}
where $\hat{A}(s_t, a_t)$ represents the annotated advantage labels. Next, we show how to integrate the fitting optimization of the advantage into the training of our PRM. For the agent tasks we attempt to solve, they have a sparse reward, which means for any time step $t < T$, $r_t = 0$ where $T$ is the final time step. As the state transition in our setting is deterministic, following previous works \citep{DBLP:journals/corr/abs-2410-08146, DBLP:journals/corr/abs-2410-11287}, we have:
\begin{align}
    Q(s_t, a_t) - V(s_t) &= Q(s_t, a_t) - (r_t + V(s_t)) \\ &= Q(s_t, a_t) - Q(s_{t-1}, a_{t-1}),
\end{align}
So the loss term for advantage $\mathcal{L}_A(\phi)$ becomes:
\begin{align}
    \nonumber
    \mathbb{E}_{s_t,a_t \sim D_Q} &\Big[\Big((\mathcal{M}_\phi(s_t,a_t)-\mathcal{M}_\phi(s_{t-1},a_{t-1})) \\
    &- (\hat{Q}(s_t,a_t) - \hat{Q}(s_{t-1},a_{t-1}))\Big)^2 \Big].
\end{align}
In this way, we can optimize our PRMs for considering not only the contribution to the final outcome but also the dependencies between adjacent actions, and our final loss for AgentPRM becomes:
\begin{equation}
    \begin{aligned}
         \mathcal{L}_{\textnormal{AgentPRM}}(\phi) = \mathcal{L_Q}(\phi) + \beta \times \mathcal{L}_A(\phi),
    \end{aligned}\label{eqn: final L_A}
\end{equation}
where $\beta$ is a scaling factor that balances the two loss terms.

\subsection{Practical Implementation for Training AgentPRM}\label{sec: Practical Implementation for Training AgentPRM}

In the previous section, we demonstrate how to optimize our AgentPRM based on the estimated or annotated data. Here, we explore how to estimate the action-value function $Q(s_t, a_t)$ in an effective and scalable manner (\textbf{\emph{RQ2}}).
We compare the previous Monte Carlo (MC) estimation, and our proposed Temporal Difference-based (TD-based) estimation with Generalized Advantage Estimation (GAE).

\subsubsection{MC-based estimation}
A common method for automatically estimating the Q-value of an action is based on Monte Carlo (MC) sampling \citep{DBLP:conf/acl/WangLSXDLCWS24, DBLP:journals/corr/abs-2406-06592}. Specifically, it first samples $N_{\textnormal{Traj}}$ seed trajectories $\{\tau_i\}_{i=1}^{N_{\textnormal{Traj}}}$ from the policy $\pi_{\theta}$. Then, for each action $a_{i,t}$ in each trajectory $\tau_i$, we start from the next state $s_{i,t+1}$ derived from the action and perform $N_{\textnormal{mc}}$ rollouts $\{\tau^{\prime}_j\}_{j=1}^{N_{\textnormal{mc}}}$ with $\pi_\theta$. As in previous work, if any of the rollouts reaches the goal state, the value of this action is set to $1$:
\begin{equation}
    \hat{Q}(s_t,a_t) =
    \begin{cases}
        1 & \exists \tau^{\prime}_j , \tau^{\prime}_j \text{ is successful,} \\
        0 & \text{otherwise,}
    \end{cases}
\end{equation}
While effective, MC-based estimation is resource-intensive and suffers from high computational cost due to the large number of rollouts required. As such, we explore alternative, more efficient methods.

\begin{figure*}[t]
    \centering
    \includegraphics[width=0.85\linewidth]{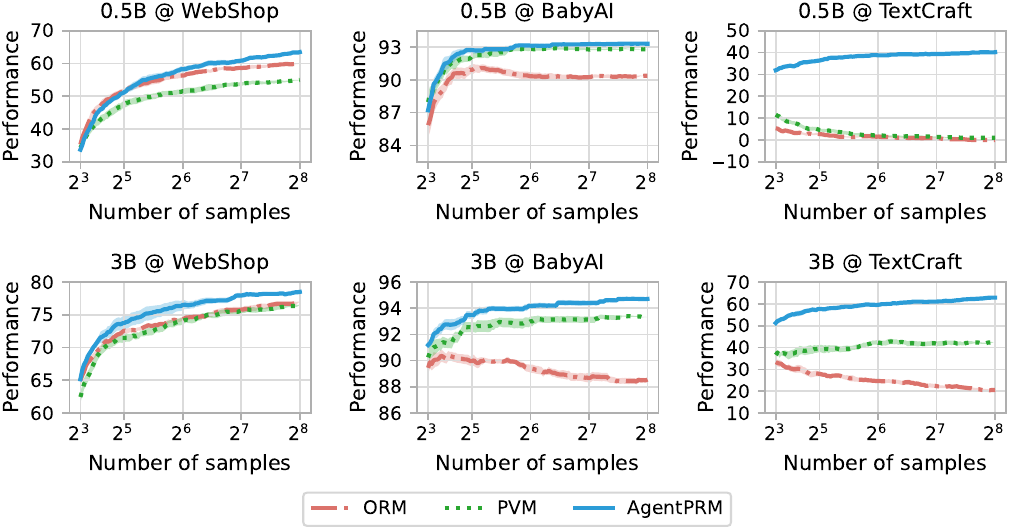}
    \caption{Performance of Best-of-N evaluation. AgentPRM outperforms other baselines, is more compute-efficient, and demonstrates a more stable and robust improvement trend as inference compute scales.}
    \label{fig: main BoN on agent tasks}
\end{figure*}

\subsubsection{TD-based estimation with GAE}
We draw inspiration from previous work \citep{DBLP:journals/corr/SchulmanMLJA15, DBLP:conf/nips/Ouyang0JAWMZASR22, sutton1999policy, sutton1988learning} and introduce TD-based methods, using GAE to reduce variance and improve stability \citep{DBLP:journals/corr/SchulmanMLJA15}. First, we define the TD residual for an action as follows:
\begin{align}
    \nonumber
    \delta (s_t, a_t) &= r_t + \gamma Q(s_t, a_t) - Q(s_{t-1}, a_{t-1}) \\
    &= r_t + \gamma \mathcal{M}_\phi(s_t, a_t) - \mathcal{M}_\phi(s_{t-1}, a_{t-1}),
\end{align}
where $r_t$ is the instant reward at timestep $t$, and in our setting, sparse rewards are only assigned when $t=T$. Next, given a trajectory $\tau = (u, o_0, a_0, o_1, \cdots, o_{T}, a_T)$, we estimate the advantage for different actions using GAE \citep{DBLP:journals/corr/SchulmanMLJA15}:
\begin{equation}
    \hat{A}(s_t, a_t) = \sum_{k=0}^{\infty} (\gamma\lambda)^{k}\delta(s_{t+k}, a_{t+k}),
\end{equation}
where $\lambda$ is the discount factor. Finally, the current estimated $\hat{Q}(s_t,a_t)$ can be represented as:
\begin{align}
    \nonumber
    \hat{Q}(s_t,a_t) &= \hat{A}(s_t,a_t) + \hat{V}(s_t) \\
    \nonumber
    &= \hat{A}(s_t,a_t) + \hat{Q}\left(s_{t-1},a_{t-1}\right) \\
    &= \hat{A}(s_t,a_t) + \mathcal{M}_\phi\left(s_{t-1},a_{t-1}\right),
\end{align}
where $t<T$. For the terminal step $t=T$, the target action-value $\hat{Q}(s_T, a_T)$ is directly defined as the final outcome reward obtained from the environment, i.e., $\hat{Q}(s_T, a_T) = r(u, \tau)$.

\begin{table*}[t]
    \centering
    \caption{Evaluation results of training-based methods and reward-model-based beam search on agent tasks. For RM-based results, @N$\times$M denotes the number of nodes retained and expanded during beam search. The best performance in each column is highlighted in bold.}
    \resizebox{0.95\linewidth}{!}{
        \begin{tabular}{ll|ccc|ccc|ccc}
            \toprule
            \multirow{1}{*}{Model} & \multirow{1}{*}{Method} & \multicolumn{3}{c|}{WebShop} & \multicolumn{3}{c|}{BabyAI} & \multicolumn{3}{c}{TextCraft} \\
            \midrule
            \multirow{7}{*}{\shortstack{Qwen2.5-0.5B}} 
            & \textit{Training-based} & & & & & & & & & \\
            & \ \ \ \ \ SFT & -- & $30.5$ & -- & -- & $37.6$ & -- & -- & $27.8$ & -- \\
            & \ \ \ \ \ RFT & -- & $39.0$ & -- & -- & $54.4$ & -- & -- & $32.9$ & --  \\
            \cdashline{2-11} 
            & \textit{Reward-Model-based} & \textit{@2$\times$2} & \textit{@4$\times$4} & \textit{@8$\times$8} &\textit{@2$\times$2} & \textit{@4$\times$4} & \textit{@8$\times$8} & \textit{@2$\times$2} & \textit{@4$\times$4} & \textit{@8$\times$8} \\
            & \ \ \ \ \ ORM & $19.5$ & $18.5$ & $8.0$ & $73.8$ & $74.9$ & $78.8$ & $25.7$ & $24.7$ & $27.8$ \\
            & \ \ \ \ \ PVM & $30.0$ & $50.0$ & $57.5$ & $\mathbf{84.6}$ & $86.5$ & $88.1$ & $25.7$ & $26.8$ & $26.8$ \\
            & \ \ \ \ \ AgentPRM & $\mathbf{30.5}$ & $\mathbf{51.5}$ & $\mathbf{62.5}$ & $82.9$ & $\mathbf{87.7}$ & $\mathbf{90.4}$ & $\mathbf{28.8}$ & $\mathbf{29.9}$ & $\mathbf{32.9}$ \\
            \midrule
            \multirow{7}{*}{\shortstack{Qwen2.5-3B}}
            &\textit{Training-based} & & & & & & & & & \\
            & \ \ \ \ \ SFT & -- & $46.0$ & -- & -- & $67.4$ & -- & -- & $29.8$ & -- \\
            & \ \ \ \ \ RFT & -- & $48.0$ & -- & -- & $64.5$ & -- & -- & $36.0$ & -- \\
            \cdashline{2-11} 
            & \textit{Reward-Model-based} & 
            \textit{@2$\times$2} & \textit{@4$\times$4} & \textit{@8$\times$8} &\textit{@2$\times$2} & \textit{@4$\times$4} & \textit{@8$\times$8} & \textit{@2$\times$2} & \textit{@4$\times$4} & \textit{@8$\times$8} \\
            & \ \ \ \ \ ORM & $51.0$ & $59.0$ & $57.0$ & $83.9$ & $83.5$ & $83.7$ & $38.1$ & $41.2$ & $43.3$ \\
            & \ \ \ \ \ PVM & $50.5$ & $59.0$ & $54.5$ & $72.7$ & $84.9$ & $89.1$ & $39.1$ & $40.2$ & $44.3$ \\
            & \ \ \ \ \ AgentPRM & $\mathbf{61.0}$ & $\mathbf{72.5}$ & $\mathbf{76.0}$ & $\mathbf{84.4}$ & $\mathbf{89.6}$ & $\mathbf{89.8}$ & $\mathbf{47.4}$ & $\mathbf{51.5}$ & $\mathbf{56.7}$ \\
            \bottomrule
        \end{tabular}
    }
    \label{tab: main beam search}
\end{table*}

In implementation, we sample $N_{\textnormal{TD}}$ trajectories $\{\tau_i\}_{i=1}^{N_\textnormal{TD}}$ from the policy $\pi_\theta$ for training. Since our estimation process involves prediction of $\mathcal{M}_\phi$, we iteratively sample a batch from the trajectory set, conduct estimation based on the current model, and update the model with Equation \ref{eqn: final L_A}. We summarize the training algorithm of AgentPRM in Algorithm \ref{algo: Training of AgentPRM}, which is also illustrated in Figure \ref{fig: AgentPRM main fig}(a).

\emph{Our method provide two main benifits}: From the efficiency perspective, TD-based estimation with GAE does not require additional rollouts from each state like MC-based method, saving a significant amount of computational resources (See \S  \ref{sec: Comparing Sampling Efficiency of Our Method with MC-based Estimation}). From the performance perspective, though TD-based methods have concerns regarding high variance, we introduce GAE to reduce variance and improve stability, ultimately achieving better performance.

\section{Experiments}\label{sec: Experiments}

\subsection{Experimental Setup}\label{subsec: Experimental Setup}

\subsubsection{Tasks.}
We conduct our experiments on three agent tasks: WebShop \citep{DBLP:conf/nips/Yao0YN22}, BabyAI \citep{DBLP:conf/iclr/Chevalier-Boisvert19}, and TextCraft \citep{DBLP:conf/naacl/PrasadKHCSBK24}.

WebShop \citep{DBLP:conf/nips/Yao0YN22}\footnote{https://github.com/princeton-nlp/WebShop/blob/master/LICENSE.md} is a simulated e-commerce website environment with 1.18 million real-world products. In this environment, an agent needs to navigate multiple types of webpages and perform diverse actions to find, customize, and purchase a product given an instruction. We set the max interaction rounds to $6$.

The BabyAI task \citep{DBLP:conf/iclr/Chevalier-Boisvert19}\footnote{https://github.com/mila-iqia/babyai/blob/master/LICENSE} involves agents navigating a grid world based on natural language instructions. The environment includes various entities such as the agent, balls, boxes, doors, and keys. Agents perform tasks like moving objects, unlocking doors, and interacting with the world according to textual commands. We set the max interaction rounds to $20$.

The TextCraft task \citep{DBLP:conf/naacl/PrasadKHCSBK24}\footnote{https://github.com/archiki/ADaPT/blob/main/LICENSE} is designed to test the ability of agents to plan and execute complex tasks that require crafting items from available resources. The dataset features a natural compositional structure, with tasks that involve a series of steps of varying complexity. The agent needs to identify and adapt to the varying task complexity. The dataset includes a variety of atomic skills, such as crafting and fetching items, and uses Minecraft's crafting recipes to specify craftable items and their ingredients. The agent's objective is to obtain target Minecraft items by crafting them from available items in the environment. We set the max interaction rounds to $20$.

Our implementation is based on AgentGym framework \citep{DBLP:journals/corr/abs-2406-04151}. We also conducted experiments on mathematical reasoning in \S  \ref{subsec: Performance on Mathematical Reasoning}.

\subsubsection{Baselines.}
We compare our AgentPRM with several training-based and reward model-based methods. For training-based approaches, supervised fine-tuning (SFT) uses expert data to fine-tune the base model, while rejection sampling fine-tuning (RFT), or self-improvement, trains the model by leveraging successful trajectories it explored \citep{DBLP:journals/corr/abs-2406-04151, zelikman2022star, yuan2023scalingrelationshiplearningmathematical, trung-etal-2024-reft, xi2024traininglargelanguagemodels}. For reward model-based methods, we include ORMs (Outcome Reward Models)\citep{DBLP:journals/corr/abs-2211-14275, DBLP:conf/nips/Ouyang0JAWMZASR22, DBLP:journals/corr/abs-2110-14168, DBLP:journals/corr/abs-2410-18451, DBLP:conf/naacl/YuGW24} and PVMs (Process Value Models) \citep{DBLP:journals/corr/abs-2211-14275, DBLP:conf/iclr/LightmanKBEBLLS24, DBLP:journals/corr/abs-2410-08146, DBLP:journals/corr/abs-2410-11287, chen2025betterprocesssupervisionbidirectional, xiong2024watchstepllmagent, xia2025agentrmenhancingagentgeneralization, miao2025boostingvirtualagentlearning, wang2025visualprmeffectiveprocessreward}. ORM estimates the reward for the outcome, while PVM estimates step-level values by assigning the reward of a trajectory to individual steps. We also include MC-based method Math-Shepherd \citep{DBLP:conf/acl/WangLSXDLCWS24} to estimate Q-Value of each step in \S  \ref{sec: Comparing Sampling Efficiency of Our Method with MC-based Estimation}.

\subsubsection{Implementation Details.}
All experiments in this work are conducted with A100-80GB GPUs. Our backbone models include Qwen-2.5-0.5B-Instruct, Qwen-2.5-3B-Instruct, Qwen-2.5-7B-Instruct \citep{qwen2.5} and Llama-3.1-8B-Instruct \citep{DBLP:journals/corr/abs-2407-21783}. For agent tasks, we use the ReAct format \citep{DBLP:conf/iclr/YaoZYDSN023} where the model first generates reasoning process and then outputs the action. To initialize the models, we randomly select $300$ trajectories from the AgentGym training set. For SFT, we set the learning rate to $1 \times 10^{-5}$. We report the success rate for WebShop and TextCraft, and the reward for BabyAI. For MC-based estimation, we set $N_\textnormal{Traj} = 1$ for each query, and $N_\textnormal{mc} = 16$ for each step; for TD-based estimation, we set $N_\textnormal{TD} = 16$ for each query. We train reward models for at most $5$ epochs under a learning rate of $1 \times 10^{-6}$. For AgentPRM, we set $\beta = 1.0$ and $\lambda = 0.95$. We set the temperature to $1.0$ in trajectory collection to maintain diversity in training data. 

\subsubsection{Evaluation Settings.}
Following AgentGym \citep{DBLP:journals/corr/abs-2406-04151}, we include $100$, $90$, $97$ queries for evaluation on WebShop \citep{DBLP:conf/nips/Yao0YN22}, BabyAI \citep{DBLP:conf/iclr/Chevalier-Boisvert19}, TextCraft \citep{DBLP:conf/naacl/PrasadKHCSBK24}, respectively. 
We perform Best-of-N and beam search to evaluate reward models as in \citep{DBLP:conf/iclr/LightmanKBEBLLS24, chen2025betterprocesssupervisionbidirectional}, setting the sampling temperature to $0.7$. For SFT and RFT method, we set the temperature to $0.0$ (i.e., greedy decoding). 

\subsection{Main Results}

\paragraph{Result 1: Compared to greedy decoding, introducing RMs for BoN and search can improve LM performance on agent tasks.}
The experimental results are shown in Figure \ref{fig: main BoN on agent tasks} and Table \ref{tab: main beam search}. Compared to the greedy decoding of SFT and RFT methods, the incorporation of reward models for Best-of-N (BoN) and search strategies significantly enhances model performance, especially as inference compute increases for more sampling. This observation aligns with prior work on test-time scaling \citep{DBLP:journals/corr/abs-2408-03314,DBLP:journals/corr/abs-2408-16737}.

\paragraph{Result 2: AgentPRM is more compute-efficient than other reward models, and outperforms them consistently in both Best-of-N and test-time search.}
As shown in Figure \ref{fig: main BoN on agent tasks}, under different sampling budgets in Best-of-N evaluation, our method consistently outperforms ORMs and PVMs across different tasks, demonstrating its effectiveness.
Figure \ref{fig: vice} (upper right) shows that, on average, AgentPRM is $8\times$ more compute-efficient than PVMs and ORMs. This highlights the potential of AgentPRM in training stronger LLM agents with methods like reinforcement learning, which we leave for future work.
As listed in Table \ref{tab: main beam search}, in beam search, our method also outperforms ORMs and PVMs across different tasks significantly, validating its ability to guide model search and achieve a good exploration-exploitation balance. For example, using Qwen2.5-3B on the WebShop task, with an $8 \times 8$ sampling search setting, our method surpasses PVM by more than $20.0$ points.

\paragraph{Result 3: As inference compute scaling, AgentPRM demonstrates a more robust and stable scaling trend.}
In Figure \ref{fig: vice} and Figure \ref{fig: main BoN on agent tasks}, we observe that as the sampling budget increases, PVMs and ORMs tend to experience performance bottleneck or even degradation. This aligns with the findings of \citet{wang2025examining}, and may be attributed to issues such as false positives or reward hacking \citep{DBLP:journals/corr/abs-2401-06080}, which could limit their effectiveness in future RL and self-improvement-based methods for training better agents. In contrast, AgentPRM consistently shows stable improvement, highlighting its robustness and broadening its potential for future applications.

\section{Discussion and Analysis}

\subsection{Ablation Study on \(\mathcal{L}_{A}(\phi)\)}\label{subsec: Ablation Study on L_A}
To capture the dependency between steps and evaluate their progress, we add $\mathcal{L}_{A}(\phi)$ for training AgentPRM. Here, we conduct an ablation on the advantage term to validate its effect. Results in Figure \ref{fig: ablation of L_A} show that without $\mathcal{L}_{A}(\phi)$, the performance on agent tasks drops regardless of the sampling strategy, showing that capturing progress is important for training AgentPRM.

\subsection{Applying AgentPRM to Reinforcement Learning}\label{sec:Applying AgentPRM to Reinforcement Learning}

\begin{figure}[htbp]
\begin{minipage}[t]{0.45\textwidth}
\includegraphics[width=\linewidth]{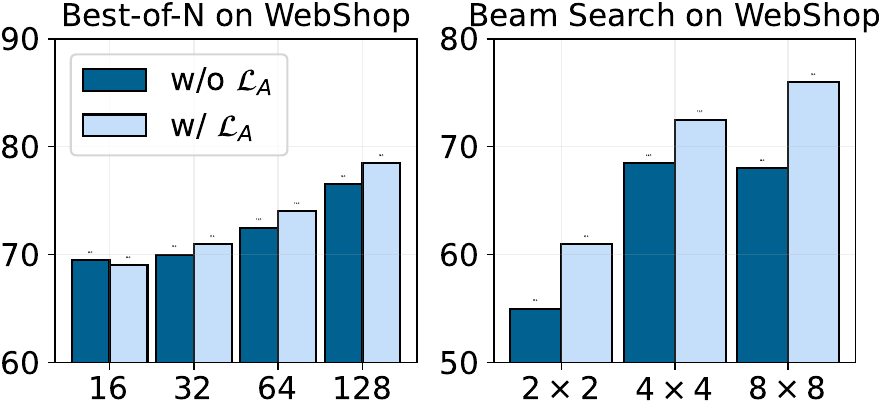}
\vspace{-10pt}
\caption{Ablation study on $\mathcal{L}_A$ with Qwen2.5-3B.}
\label{fig: ablation of L_A}
\vspace{-10pt}
\end{minipage} 
\hfill 
\begin{minipage}[t]{0.5\textwidth}
\includegraphics[width=\linewidth]{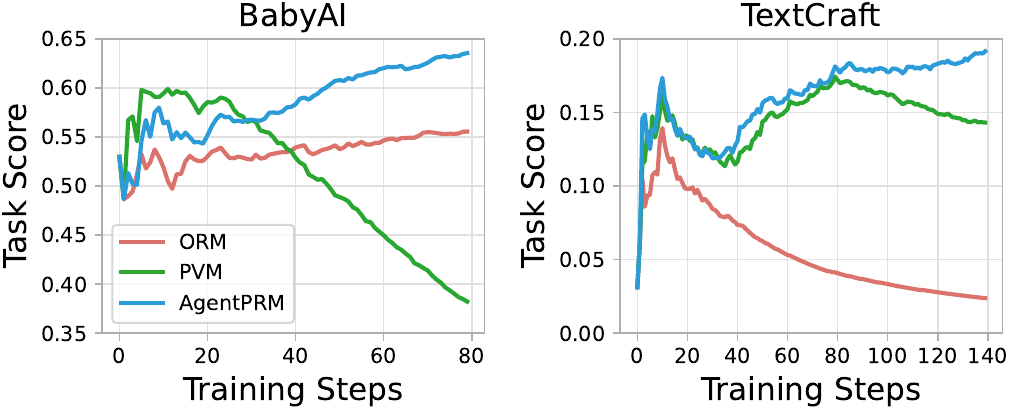} 
\vspace{-10pt}
\caption{Task score of RL optimization.}
\label{fig: RL}
\end{minipage}
\end{figure}

Reinforcement learning (RL) has become a core method for training LLMs \citep{DBLP:conf/nips/Ouyang0JAWMZASR22, openai2024o1, deepseekai2025deepseekr1incentivizingreasoningcapability}. To assess the effectiveness of our reward model, we integrate it into the RL optimization process. We conduct experiments with Qwen2.5-3B on BabyAI and TextCraft, using Proximal Policy Optimization (PPO) as the algorithm as it is widely adopted. 
More implementation details are in Appendix \ref{app: Implementation Details of Reinforcement Learning}.

As shown in Figure \ref{fig: RL}, while other baseline reward models face optimization instability or slower performance improvements, AgentPRM delivers a more stable and effective optimization, outperforming the baselines and highlighting its advantages. 

\subsection{Performance on Mathematical Reasoning}\label{subsec: Performance on Mathematical Reasoning}

\begin{figure}[h] 
\begin{minipage}[c]{0.42\textwidth}
\centering
\includegraphics[width=\linewidth, height=3.5cm, keepaspectratio]{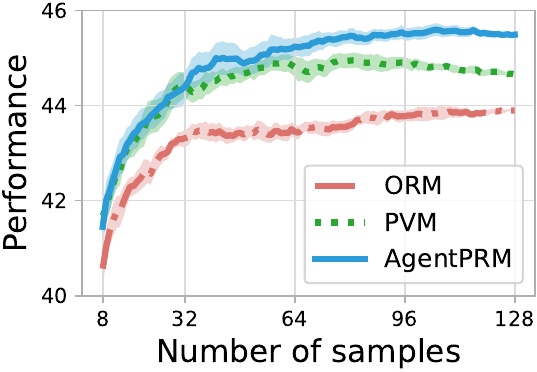}
\caption{Best-of-N results on mathematical reasoning tasks.}    
\label{fig: BoN on GSM8K}
\end{minipage}%
\hfill
\begin{minipage}[c]{0.55\textwidth}
\captionof{table}{Evaluation results of beam search on mathematical reasoning tasks.}  
\footnotesize 
\label{tab: beam search on GSM8K}
\vspace{0.2cm}  
\resizebox{!}{2.0cm}{  
    \begin{tabular}{llccc}
        \toprule
        \multirow{2}{*}{Model} & \multirow{2}{*}{Method} & \multicolumn{3}{c}{Math Reasoning} \\
        \cmidrule(lr){3-5} 
        & & $@2\times2$ & $@4\times4$ & $@8\times8$ \\
        \midrule
        \multirowcell{3}[0pt][l]{Qwen2.5-0.5B} & ORM & $39.5$ & $42.9$ & $44.2$ \\
        & PVM & $38.9$ & $41.3$ & $42.9$ \\
        & AgentPRM & $\mathbf{41.5}$ & $\mathbf{44.7}$ & $\mathbf{45.7}$ \\
        \midrule
        \multirowcell{3}[0pt][l]{Qwen2.5-3B} & ORM & $64.1$ & $70.3$ & $72.9$ \\
        & PVM & $63.6$ & $69.6$ & $71.2$ \\
        & AgentPRM & $\mathbf{65.1}$ & $\mathbf{70.5}$ & $\mathbf{73.4}$ \\
        \bottomrule
    \end{tabular}
}
\end{minipage}  
\end{figure}

To demonstrate the versatility of our method, we also conducted experiments on mathematical reasoning tasks. 
Following Math-Shepherd \citep{DBLP:conf/acl/WangLSXDLCWS24}, we formulate math reasoning as multi-turn decision making: at step $t$, the partial solution is defined as the state $s_t$ and the next reasoning step is the action $a_t$. The model emits stepwise text with a delimiter token after each step to segment steps.
Since there is no external environmental feedback in this setting, generation simply resumes from the end of the previous step. This yields a well-defined multi-step generation protocol for mathematical reasoning.

In experiments, we employ the GSM8K dataset \citep{DBLP:journals/corr/abs-2110-14168}, with results shown in Figure \ref{fig: BoN on GSM8K} and Table \ref{tab: beam search on GSM8K}. 
As we can see, our method still performs exceptionally well on mathematical reasoning tasks, surpassing other baselines. This also highlights the generalizability and adaptability of our AgentPRM. We expect to extend it to more tasks in future work, such as coding or logical reasoning.

\subsection{Comparing Sampling Efficiency of Our Method with MC-based Estimation}\label{sec: Comparing Sampling Efficiency of Our Method with MC-based Estimation}
\begin{table}[h]
    \centering
    \caption{Sampling cost and performance of our method and MC-based estimation for PRMs. “Tokens” denotes the amount of sampled tokens used to train PRMs.}
    \begin{tabular}{lllcccccc}
        \toprule
        \multirow{2}{*}{Task} & \multirow{2}{*}{Method} & \multirow{2}{*}{Tokens} & \multicolumn{4}{c}{Best-of-N} & \multicolumn{2}{c}{Beam Search} \\
        \cmidrule(lr){4-7} \cmidrule(lr){8-9}
        & & & $@8$ & $@16$ & $@32$ & $@64$ & $@4\times4$ & $@8\times8$ \\
        \midrule
        \multirowcell{2}[0pt][l]{WebShop} & MC-based & $1.9\times$ & $63.5$ & $67.5$ & $69.0$ & $72.0$ & $67.5$ & $70.5$ \\
        & TD-based & $1.0\times$ & $\mathbf{64.5}$ & $\mathbf{69.0}$ & $\mathbf{71.0}$ & $\mathbf{74.0}$ & $\mathbf{72.5}$ & $\mathbf{76.0}$ \\
        \midrule
        \multirowcell{2}[0pt][l]{BabyAI} & MC-based & $2.8\times$ & $90.5$ & $90.5$ & $91.6$ & $93.1$ & $87.6$ & $88.3$ \\
        & TD-based & $1.0\times$ & $\mathbf{91.4}$ & $\mathbf{91.4}$ & $\mathbf{92.4}$ & $\mathbf{94.4}$ & $\mathbf{89.6}$ & $\mathbf{89.8}$ \\
        \midrule
        \multirowcell{2}[0pt][l]{Math} & MC-based & $1.5\times$ & $61.8$ & $63.3$ & $63.9$ & $65.0$ & $66.1$ & $70.1$ \\
        & TD-based & $1.0\times$ & $\mathbf{68.9}$ & $\mathbf{72.4}$ & $\mathbf{73.9}$ & $\mathbf{74.7}$ & $\mathbf{70.5}$ & $\mathbf{73.4}$ \\
        \bottomrule
    \end{tabular}
    \label{tab: sampling efficiency}
\end{table}

In \S  \ref{sec: Practical Implementation for Training AgentPRM}, we introduced TD-based estimation with GAE for the automated labeling process and stated its benifits. Here, we empirically compare it with the previously commonly used MC-based estimation. The results are shown in Table \ref{tab: sampling efficiency}. We observe that our method requires fewer tokens for labeling the data (i.e., more data-efficient) compared to other methods, yet achieves better performance on Best-of-N and beam search, demonstrating the higher efficiency and effectiveness of our approach.

\subsection{Evaluating Value Distributions of Actions with AgentPRM}
\begin{figure}[ht] 

\begin{minipage}[t]{0.5\textwidth}
\includegraphics[width=0.48\linewidth]{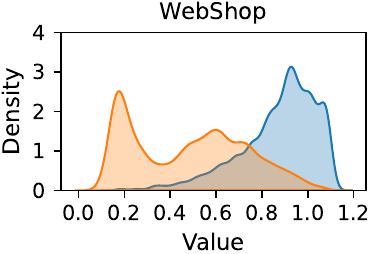}
\includegraphics[width=0.48\linewidth]{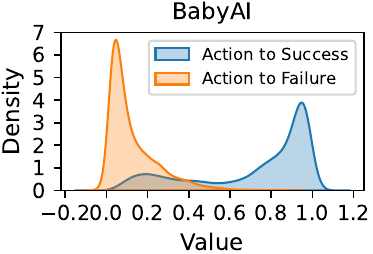}
\caption{Visualization of value distribution of \\ Actions with AgentPRM.}
\label{fig: value distribution}
\end{minipage} 
\hspace{0.01\textwidth} 
\begin{minipage}[t]{0.5\textwidth}
\includegraphics[width=0.45\linewidth]{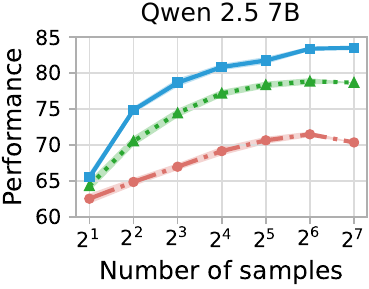}
\includegraphics[width=0.45\linewidth]{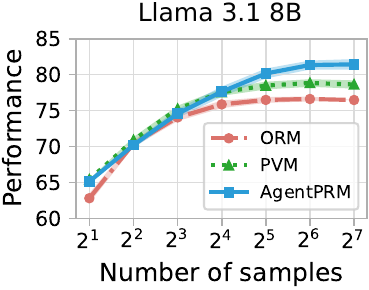}
\caption{Average BoN performance on Qwen2.5-7B and Llama3.1-8B across three tasks.}
\label{fig: more BoN}
 \end{minipage}   
\end{figure}

To further demonstrate the working mechanism of AgentPRM, we visualize the value estimates of the actions predicted by AgentPRM in WebShop and BabyAI across successful and unsuccessful trajectories. From the distribution in Figure \ref{fig: value distribution}, we observe that the model assigns higher scores to the actions that lead to positive goals, and lower scores to the actions that lead to negative goals, revealing that our method is effective in credit assignment.

\subsection{Experiments on Models of Larger Size and Different Series}

We also validate the effectiveness of AgentPRM on larger Qwen-2.5-7B-Instruct and the Llama-series (i.e., Llama3.1-8B-Instruct). Figure \ref{fig: more BoN} illustrates the average results across three agentic tasks (i.e., WebShop, BabyAI, and TextCraft). We can find that AgentPRM consistently outperforms the baseline methods on the two models, demonstrating its generalization across model sizes and architectures.

We also perform a qualitative analysis in Appendix \ref{app: Qualitative Analysis} to show how AgentPRM works.

\section{Related Work}\label{sec: Related Works}

\paragraph{Developing LLMs for agent tasks.}
To enable language models to perform well in multi-turn decision-making tasks \citep{DBLP:conf/nips/Yao0YN22, DBLP:conf/iclr/ZhouX0ZLSCOBF0N24, DBLP:conf/iclr/Chevalier-Boisvert19}, previous work has proposed training-based methods, where expert-labeled trajectories are collected, and the learner imitates them step by step \citep{DBLP:conf/acl/ChenLWZLLCZ24, DBLP:journals/corr/abs-2310-05915}. However, this approach is often difficult to scale and lacks sufficient exploration of the environment by the model. Prompt-engineering-based methods leverage SOTA commercial models like GPT-4o for developing agents, which is limited by APIs, making it difficult to customize \citep{DBLP:journals/corr/abs-2306-02224, DBLP:journals/corr/abs-2407-01476}. Another line of work adopts self-improvement methods \citep{DBLP:journals/corr/abs-2406-04151, aksitov2023restmeetsreactselfimprovement, song2024trialerrorexplorationbasedtrajectory, yang2024reactmeetsactrelanguage, tao2024surveyselfevolutionlargelanguage}, allowing the model to explore and learn within the environment. However, these approaches typically rely solely on outcome-based feedback and fail to assess the value and impact of each individual decision \citep{ARMAP2024, QAgent2025}. In this paper, we explore training PRMs to guide the exploration of LMs, decoupling it from the optimization of the agent, and the resulted PRMs can also be used as verifiers for re-ranking and search.

\paragraph{PRMs for LLMs.}
PRMs can provide dense reward signals to help LLMs in RL and test-time search or re-ranking \citep{DBLP:journals/corr/abs-2408-03314}, and are widely used in LLM reasoning \citep{DBLP:conf/acl/WangLSXDLCWS24, DBLP:conf/iclr/LightmanKBEBLLS24, DBLP:journals/corr/abs-2410-11287, DBLP:conf/naacl/YuGW24}. However, the data labeling required for this approach is expensive and not scalable \citep{DBLP:conf/iclr/LightmanKBEBLLS24}. Therefore, recent work has explored automated annotating methods based on Monte Carlo sampling to reduce the cost \citep{DBLP:conf/acl/WangLSXDLCWS24, DBLP:journals/corr/abs-2410-11287}. In agent tasks, some works have also used similar MC sampling methods to label the Q-values of actions \citep{QAgent2025, DBLP:conf/emnlp/HaoGMHWWH23, DBLP:journals/corr/abs-2409-09345}. However, they only consider the future success probability of a step, without accounting for the dependencies and progress between steps \citep{DBLP:journals/corr/abs-2309-07864, DBLP:journals/corr/abs-2406-04151, DBLP:conf/nips/Yao0YN22, DBLP:conf/iclr/Chevalier-Boisvert19}. Our AgentPRM captures both of these aspects and we perform data labeling more efficiently by using the method of TD-estimation with GAE.

See Appendix \ref{app: More Detailed Discussion of Related Work} for more detailed discussion of related work.

\section{Conclusion}

In this paper, we introduce AgentPRM, a process supervision model designed for LLM agents in multi-step decision-making tasks. It captures both the probability of each step achieving the goal (promise) and the interdependence between sequential steps (progress). Extensive experiments demonstrate that our method outperforms other baselines across various sampling strategies, models, and tasks. Additionally, it is more compute-efficient, and its performance shows robust improvement as inference compute increases, highlighting its potential for training stronger agents in the future. Moreover, our method generalizes well to mathematical tasks, showcasing its versatility. We also conducted extensive additional analyses and ablation to demonstrate how our method works, its data efficiency, and its adaptability to different model architectures and sizes. We hope our work can provide valuable insights and contributions for the LLM agent community.

\bibliography{main}

@misc{DBLP:conf/nips/Yao0YN22,
  author       = {Shunyu Yao and
                  Howard Chen and
                  John Yang and
                  Karthik Narasimhan},
  editor       = {Sanmi Koyejo and
                  S. Mohamed and
                  A. Agarwal and
                  Danielle Belgrave and
                  K. Cho and
                  A. Oh},
  title        = {WebShop: Towards Scalable Real-World Web Interaction with Grounded
                  Language Agents},
  booktitle    = {Advances in Neural Information Processing Systems 35: Annual Conference
                  on Neural Information Processing Systems 2022, NeurIPS 2022, New Orleans,
                  LA, USA, November 28 - December 9, 2022},
  year         = {2022},
  url          = {http://papers.nips.cc/paper\_files/paper/2022/hash/82ad13ec01f9fe44c01cb91814fd7b8c-Abstract-Conference.html},
  bibsource    = {dblp computer science bibliography, https://dblp.org}
}

@misc{DBLP:conf/iclr/ZhouX0ZLSCOBF0N24,
  author       = {Shuyan Zhou and
                  Frank F. Xu and
                  Hao Zhu and
                  Xuhui Zhou and
                  Robert Lo and
                  Abishek Sridhar and
                  Xianyi Cheng and
                  Tianyue Ou and
                  Yonatan Bisk and
                  Daniel Fried and
                  Uri Alon and
                  Graham Neubig},
  title        = {WebArena: {A} Realistic Web Environment for Building Autonomous Agents},
  booktitle    = {The Twelfth International Conference on Learning Representations,
                  {ICLR} 2024, Vienna, Austria, May 7-11, 2024},
  publisher    = {OpenReview.net},
  year         = {2024},
  url          = {https://openreview.net/forum?id=oKn9c6ytLx},
  bibsource    = {dblp computer science bibliography, https://dblp.org}
}

@misc{DBLP:journals/corr/abs-2307-09288,
  author       = {Hugo Touvron and
                  Louis Martin and
                  Kevin Stone and
                  Peter Albert and
                  Amjad Almahairi and
                  Yasmine Babaei and
                  Nikolay Bashlykov and
                  Soumya Batra and
                  Prajjwal Bhargava and
                  Shruti Bhosale and
                  Dan Bikel and
                  Lukas Blecher and
                  Cristian Canton{-}Ferrer and
                  Moya Chen and
                  Guillem Cucurull and
                  David Esiobu and
                  Jude Fernandes and
                  Jeremy Fu and
                  Wenyin Fu and
                  Brian Fuller and
                  Cynthia Gao and
                  Vedanuj Goswami and
                  Naman Goyal and
                  Anthony Hartshorn and
                  Saghar Hosseini and
                  Rui Hou and
                  Hakan Inan and
                  Marcin Kardas and
                  Viktor Kerkez and
                  Madian Khabsa and
                  Isabel Kloumann and
                  Artem Korenev and
                  Punit Singh Koura and
                  Marie{-}Anne Lachaux and
                  Thibaut Lavril and
                  Jenya Lee and
                  Diana Liskovich and
                  Yinghai Lu and
                  Yuning Mao and
                  Xavier Martinet and
                  Todor Mihaylov and
                  Pushkar Mishra and
                  Igor Molybog and
                  Yixin Nie and
                  Andrew Poulton and
                  Jeremy Reizenstein and
                  Rashi Rungta and
                  Kalyan Saladi and
                  Alan Schelten and
                  Ruan Silva and
                  Eric Michael Smith and
                  Ranjan Subramanian and
                  Xiaoqing Ellen Tan and
                  Binh Tang and
                  Ross Taylor and
                  Adina Williams and
                  Jian Xiang Kuan and
                  Puxin Xu and
                  Zheng Yan and
                  Iliyan Zarov and
                  Yuchen Zhang and
                  Angela Fan and
                  Melanie Kambadur and
                  Sharan Narang and
                  Aur{\'{e}}lien Rodriguez and
                  Robert Stojnic and
                  Sergey Edunov and
                  Thomas Scialom},
  title        = {Llama 2: Open Foundation and Fine-Tuned Chat Models},
  journal      = {CoRR},
  volume       = {abs/2307.09288},
  year         = {2023},
  url          = {https://doi.org/10.48550/arXiv.2307.09288},
  doi          = {10.48550/ARXIV.2307.09288},
  eprinttype    = {arXiv},
  eprint       = {2307.09288},
  bibsource    = {dblp computer science bibliography, https://dblp.org}
}

@misc{DBLP:journals/corr/abs-2407-21783,
  author       = {Abhimanyu Dubey and
                  Abhinav Jauhri and
                  Abhinav Pandey and
                  Abhishek Kadian and
                  Ahmad Al{-}Dahle and
                  Aiesha Letman and
                  Akhil Mathur and
                  Alan Schelten and
                  Amy Yang and
                  Angela Fan and
                  Anirudh Goyal and
                  Anthony Hartshorn and
                  Aobo Yang and
                  Archi Mitra and
                  Archie Sravankumar and
                  Artem Korenev and
                  Arthur Hinsvark and
                  Arun Rao and
                  Aston Zhang and
                  Aur{\'{e}}lien Rodriguez and
                  Austen Gregerson and
                  Ava Spataru and
                  Baptiste Rozi{\`{e}}re and
                  Bethany Biron and
                  Binh Tang and
                  Bobbie Chern and
                  Charlotte Caucheteux and
                  Chaya Nayak and
                  Chloe Bi and
                  Chris Marra and
                  Chris McConnell and
                  Christian Keller and
                  Christophe Touret and
                  Chunyang Wu and
                  Corinne Wong and
                  Cristian Canton Ferrer and
                  Cyrus Nikolaidis and
                  Damien Allonsius and
                  Daniel Song and
                  Danielle Pintz and
                  Danny Livshits and
                  David Esiobu and
                  Dhruv Choudhary and
                  Dhruv Mahajan and
                  Diego Garcia{-}Olano and
                  Diego Perino and
                  Dieuwke Hupkes and
                  Egor Lakomkin and
                  Ehab AlBadawy and
                  Elina Lobanova and
                  Emily Dinan and
                  Eric Michael Smith and
                  Filip Radenovic and
                  Frank Zhang and
                  Gabriel Synnaeve and
                  Gabrielle Lee and
                  Georgia Lewis Anderson and
                  Graeme Nail and
                  Gr{\'{e}}goire Mialon and
                  Guan Pang and
                  Guillem Cucurell and
                  Hailey Nguyen and
                  Hannah Korevaar and
                  Hu Xu and
                  Hugo Touvron and
                  Iliyan Zarov and
                  Imanol Arrieta Ibarra and
                  Isabel M. Kloumann and
                  Ishan Misra and
                  Ivan Evtimov and
                  Jade Copet and
                  Jaewon Lee and
                  Jan Geffert and
                  Jana Vranes and
                  Jason Park and
                  Jay Mahadeokar and
                  Jeet Shah and
                  Jelmer van der Linde and
                  Jennifer Billock and
                  Jenny Hong and
                  Jenya Lee and
                  Jeremy Fu and
                  Jianfeng Chi and
                  Jianyu Huang and
                  Jiawen Liu and
                  Jie Wang and
                  Jiecao Yu and
                  Joanna Bitton and
                  Joe Spisak and
                  Jongsoo Park and
                  Joseph Rocca and
                  Joshua Johnstun and
                  Joshua Saxe and
                  Junteng Jia and
                  Kalyan Vasuden Alwala and
                  Kartikeya Upasani and
                  Kate Plawiak and
                  Ke Li and
                  Kenneth Heafield and
                  Kevin Stone and
                  et al.},
  title        = {The Llama 3 Herd of Models},
  journal      = {CoRR},
  volume       = {abs/2407.21783},
  year         = {2024},
  url          = {https://doi.org/10.48550/arXiv.2407.21783},
  doi          = {10.48550/ARXIV.2407.21783},
  eprinttype    = {arXiv},
  eprint       = {2407.21783},
  bibsource    = {dblp computer science bibliography, https://dblp.org}
}

@misc{openai2024GPT4o,
  author       = {OpenAI},
  title        = {Hello GPT-4o},
  year         = {2024},
  month        = {5},
  url          = {https://openai.com/index/hello-gpt-4o/},
}

@misc{qwen2.5,
    title = {Qwen2.5: A Party of Foundation Models},
    url = {https://qwenlm.github.io/blog/qwen2.5/},
    author = {QwenTeam},
    month = {September},
    year = {2024}
}

@misc{DBLP:journals/corr/abs-2312-11805,
  author       = {Rohan Anil and
                  Sebastian Borgeaud and
                  Yonghui Wu and
                  Jean{-}Baptiste Alayrac and
                  Jiahui Yu and
                  Radu Soricut and
                  Johan Schalkwyk and
                  Andrew M. Dai and
                  Anja Hauth and
                  Katie Millican and
                  David Silver and
                  Slav Petrov and
                  Melvin Johnson and
                  Ioannis Antonoglou and
                  Julian Schrittwieser and
                  Amelia Glaese and
                  Jilin Chen and
                  Emily Pitler and
                  Timothy P. Lillicrap and
                  Angeliki Lazaridou and
                  Orhan Firat and
                  James Molloy and
                  Michael Isard and
                  Paul Ronald Barham and
                  Tom Hennigan and
                  Benjamin Lee and
                  Fabio Viola and
                  Malcolm Reynolds and
                  Yuanzhong Xu and
                  Ryan Doherty and
                  Eli Collins and
                  Clemens Meyer and
                  Eliza Rutherford and
                  Erica Moreira and
                  Kareem Ayoub and
                  Megha Goel and
                  George Tucker and
                  Enrique Piqueras and
                  Maxim Krikun and
                  Iain Barr and
                  Nikolay Savinov and
                  Ivo Danihelka and
                  Becca Roelofs and
                  Ana{\"{\i}}s White and
                  Anders Andreassen and
                  Tamara von Glehn and
                  Lakshman Yagati and
                  Mehran Kazemi and
                  Lucas Gonzalez and
                  Misha Khalman and
                  Jakub Sygnowski and
                  et al.},
  title        = {Gemini: {A} Family of Highly Capable Multimodal Models},
  journal      = {CoRR},
  volume       = {abs/2312.11805},
  year         = {2023},
  url          = {https://doi.org/10.48550/arXiv.2312.11805},
  doi          = {10.48550/ARXIV.2312.11805},
  eprinttype    = {arXiv},
  eprint       = {2312.11805},
  bibsource    = {dblp computer science bibliography, https://dblp.org}
}

@misc{DBLP:conf/acl/ZengLLWLD024,
  author       = {Aohan Zeng and
                  Mingdao Liu and
                  Rui Lu and
                  Bowen Wang and
                  Xiao Liu and
                  Yuxiao Dong and
                  Jie Tang},
  editor       = {Lun{-}Wei Ku and
                  Andre Martins and
                  Vivek Srikumar},
  title        = {AgentTuning: Enabling Generalized Agent Abilities for LLMs},
  booktitle    = {Findings of the Association for Computational Linguistics, {ACL} 2024,
                  Bangkok, Thailand and virtual meeting, August 11-16, 2024},
  pages        = {3053--3077},
  publisher    = {Association for Computational Linguistics},
  year         = {2024},
  url          = {https://doi.org/10.18653/v1/2024.findings-acl.181},
  doi          = {10.18653/V1/2024.FINDINGS-ACL.181},
  bibsource    = {dblp computer science bibliography, https://dblp.org}
}

@misc{DBLP:conf/acl/ChenLWZLLCZ24,
  author       = {Zehui Chen and
                  Kuikun Liu and
                  Qiuchen Wang and
                  Wenwei Zhang and
                  Jiangning Liu and
                  Dahua Lin and
                  Kai Chen and
                  Feng Zhao},
  editor       = {Lun{-}Wei Ku and
                  Andre Martins and
                  Vivek Srikumar},
  title        = {Agent-FLAN: Designing Data and Methods of Effective Agent Tuning for
                  Large Language Models},
  booktitle    = {Findings of the Association for Computational Linguistics, {ACL} 2024,
                  Bangkok, Thailand and virtual meeting, August 11-16, 2024},
  pages        = {9354--9366},
  publisher    = {Association for Computational Linguistics},
  year         = {2024},
  url          = {https://doi.org/10.18653/v1/2024.findings-acl.557},
  doi          = {10.18653/V1/2024.FINDINGS-ACL.557},
  bibsource    = {dblp computer science bibliography, https://dblp.org}
}

@misc{DBLP:journals/corr/abs-2406-04151,
  author       = {Zhiheng Xi and
                  Yiwen Ding and
                  Wenxiang Chen and
                  Boyang Hong and
                  Honglin Guo and
                  Junzhe Wang and
                  Dingwen Yang and
                  Chenyang Liao and
                  Xin Guo and
                  Wei He and
                  Songyang Gao and
                  Lu Chen and
                  Rui Zheng and
                  Yicheng Zou and
                  Tao Gui and
                  Qi Zhang and
                  Xipeng Qiu and
                  Xuanjing Huang and
                  Zuxuan Wu and
                  Yu{-}Gang Jiang},
  title        = {AgentGym: Evolving Large Language Model-based Agents across Diverse
                  Environments},
  journal      = {CoRR},
  volume       = {abs/2406.04151},
  year         = {2024},
  url          = {https://doi.org/10.48550/arXiv.2406.04151},
  doi          = {10.48550/ARXIV.2406.04151},
  eprinttype    = {arXiv},
  eprint       = {2406.04151},
  bibsource    = {dblp computer science bibliography, https://dblp.org}
}

@misc{DBLP:journals/corr/abs-2308-05960,
  author       = {Zhiwei Liu and
                  Weiran Yao and
                  Jianguo Zhang and
                  Le Xue and
                  Shelby Heinecke and
                  Rithesh Murthy and
                  Yihao Feng and
                  Zeyuan Chen and
                  Juan Carlos Niebles and
                  Devansh Arpit and
                  Ran Xu and
                  Phil Mui and
                  Huan Wang and
                  Caiming Xiong and
                  Silvio Savarese},
  title        = {{BOLAA:} Benchmarking and Orchestrating LLM-augmented Autonomous Agents},
  journal      = {CoRR},
  volume       = {abs/2308.05960},
  year         = {2023},
  url          = {https://doi.org/10.48550/arXiv.2308.05960},
  doi          = {10.48550/ARXIV.2308.05960},
  eprinttype    = {arXiv},
  eprint       = {2308.05960},
  bibsource    = {dblp computer science bibliography, https://dblp.org}
}

@misc{DBLP:conf/nips/ShinnCGNY23,
  author       = {Noah Shinn and
                  Federico Cassano and
                  Ashwin Gopinath and
                  Karthik Narasimhan and
                  Shunyu Yao},
  editor       = {Alice Oh and
                  Tristan Naumann and
                  Amir Globerson and
                  Kate Saenko and
                  Moritz Hardt and
                  Sergey Levine},
  title        = {Reflexion: language agents with verbal reinforcement learning},
  booktitle    = {Advances in Neural Information Processing Systems 36: Annual Conference
                  on Neural Information Processing Systems 2023, NeurIPS 2023, New Orleans,
                  LA, USA, December 10 - 16, 2023},
  year         = {2023},
  url          = {http://papers.nips.cc/paper\_files/paper/2023/hash/1b44b878bb782e6954cd888628510e90-Abstract-Conference.html},
  bibsource    = {dblp computer science bibliography, https://dblp.org}
}

@misc{DBLP:conf/acl/WangLSXDLCWS24,
  author       = {Peiyi Wang and
                  Lei Li and
                  Zhihong Shao and
                  Runxin Xu and
                  Damai Dai and
                  Yifei Li and
                  Deli Chen and
                  Yu Wu and
                  Zhifang Sui},
  editor       = {Lun{-}Wei Ku and
                  Andre Martins and
                  Vivek Srikumar},
  title        = {Math-Shepherd: Verify and Reinforce LLMs Step-by-step without Human
                  Annotations},
  booktitle    = {Proceedings of the 62nd Annual Meeting of the Association for Computational
                  Linguistics (Volume 1: Long Papers), {ACL} 2024, Bangkok, Thailand,
                  August 11-16, 2024},
  pages        = {9426--9439},
  publisher    = {Association for Computational Linguistics},
  year         = {2024},
  url          = {https://doi.org/10.18653/v1/2024.acl-long.510},
  doi          = {10.18653/V1/2024.ACL-LONG.510},
  bibsource    = {dblp computer science bibliography, https://dblp.org}
}

@misc{DBLP:journals/fcsc/WangMFZYZCTCLZWW24,
  author       = {Lei Wang and
                  Chen Ma and
                  Xueyang Feng and
                  Zeyu Zhang and
                  Hao Yang and
                  Jingsen Zhang and
                  Zhiyuan Chen and
                  Jiakai Tang and
                  Xu Chen and
                  Yankai Lin and
                  Wayne Xin Zhao and
                  Zhewei Wei and
                  Jirong Wen},
  title        = {A survey on large language model based autonomous agents},
  journal      = {Frontiers Comput. Sci.},
  volume       = {18},
  number       = {6},
  pages        = {186345},
  year         = {2024},
  url          = {https://doi.org/10.1007/s11704-024-40231-1},
  doi          = {10.1007/S11704-024-40231-1},
  bibsource    = {dblp computer science bibliography, https://dblp.org}
}

@misc{DBLP:journals/corr/SchulmanMLJA15,
  author       = {John Schulman and
                  Philipp Moritz and
                  Sergey Levine and
                  Michael I. Jordan and
                  Pieter Abbeel},
  editor       = {Yoshua Bengio and
                  Yann LeCun},
  title        = {High-Dimensional Continuous Control Using Generalized Advantage Estimation},
  booktitle    = {4th International Conference on Learning Representations, {ICLR} 2016,
                  San Juan, Puerto Rico, May 2-4, 2016, Conference Track Proceedings},
  year         = {2016},
  url          = {http://arxiv.org/abs/1506.02438},
  bibsource    = {dblp computer science bibliography, https://dblp.org}
}

@misc{DBLP:conf/aaaifs/HausknechtS15,
  author       = {Matthew J. Hausknecht and
                  Peter Stone},
  title        = {Deep Recurrent Q-Learning for Partially Observable MDPs},
  booktitle    = {2015 {AAAI} Fall Symposia, Arlington, Virginia, USA, November 12-14,
                  2015},
  pages        = {29--37},
  publisher    = {{AAAI} Press},
  year         = {2015},
  url          = {http://www.aaai.org/ocs/index.php/FSS/FSS15/paper/view/11673},
  bibsource    = {dblp computer science bibliography, https://dblp.org}
}

@misc{DBLP:journals/corr/abs-2211-14275,
  author       = {Jonathan Uesato and
                  Nate Kushman and
                  Ramana Kumar and
                  H. Francis Song and
                  Noah Y. Siegel and
                  Lisa Wang and
                  Antonia Creswell and
                  Geoffrey Irving and
                  Irina Higgins},
  title        = {Solving math word problems with process- and outcome-based feedback},
  journal      = {CoRR},
  volume       = {abs/2211.14275},
  year         = {2022},
  url          = {https://doi.org/10.48550/arXiv.2211.14275},
  doi          = {10.48550/ARXIV.2211.14275},
  eprinttype    = {arXiv},
  eprint       = {2211.14275},
  bibsource    = {dblp computer science bibliography, https://dblp.org}
}

@misc{DBLP:conf/nips/Ouyang0JAWMZASR22,
  author       = {Long Ouyang and
                  Jeffrey Wu and
                  Xu Jiang and
                  Diogo Almeida and
                  Carroll L. Wainwright and
                  Pamela Mishkin and
                  Chong Zhang and
                  Sandhini Agarwal and
                  Katarina Slama and
                  Alex Ray and
                  John Schulman and
                  Jacob Hilton and
                  Fraser Kelton and
                  Luke Miller and
                  Maddie Simens and
                  Amanda Askell and
                  Peter Welinder and
                  Paul F. Christiano and
                  Jan Leike and
                  Ryan Lowe},
  editor       = {Sanmi Koyejo and
                  S. Mohamed and
                  A. Agarwal and
                  Danielle Belgrave and
                  K. Cho and
                  A. Oh},
  title        = {Training language models to follow instructions with human feedback},
  booktitle    = {Advances in Neural Information Processing Systems 35: Annual Conference
                  on Neural Information Processing Systems 2022, NeurIPS 2022, New Orleans,
                  LA, USA, November 28 - December 9, 2022},
  year         = {2022},
  url          = {http://papers.nips.cc/paper\_files/paper/2022/hash/b1efde53be364a73914f58805a001731-Abstract-Conference.html},
  bibsource    = {dblp computer science bibliography, https://dblp.org}
}

@misc{DBLP:journals/corr/abs-2410-18451,
  author       = {Chris Yuhao Liu and
                  Liang Zeng and
                  Jiacai Liu and
                  Rui Yan and
                  Jujie He and
                  Chaojie Wang and
                  Shuicheng Yan and
                  Yang Liu and
                  Yahui Zhou},
  title        = {Skywork-Reward: Bag of Tricks for Reward Modeling in LLMs},
  journal      = {CoRR},
  volume       = {abs/2410.18451},
  year         = {2024},
  url          = {https://doi.org/10.48550/arXiv.2410.18451},
  doi          = {10.48550/ARXIV.2410.18451},
  eprinttype    = {arXiv},
  eprint       = {2410.18451},
  bibsource    = {dblp computer science bibliography, https://dblp.org}
}

@misc{DBLP:journals/corr/abs-2110-14168,
  author       = {Karl Cobbe and
                  Vineet Kosaraju and
                  Mohammad Bavarian and
                  Mark Chen and
                  Heewoo Jun and
                  Lukasz Kaiser and
                  Matthias Plappert and
                  Jerry Tworek and
                  Jacob Hilton and
                  Reiichiro Nakano and
                  Christopher Hesse and
                  John Schulman},
  title        = {Training Verifiers to Solve Math Word Problems},
  journal      = {CoRR},
  volume       = {abs/2110.14168},
  year         = {2021},
  url          = {https://arxiv.org/abs/2110.14168},
  eprinttype    = {arXiv},
  eprint       = {2110.14168},
  bibsource    = {dblp computer science bibliography, https://dblp.org}
}

@misc{DBLP:conf/iclr/LightmanKBEBLLS24,
  author       = {Hunter Lightman and
                  Vineet Kosaraju and
                  Yuri Burda and
                  Harrison Edwards and
                  Bowen Baker and
                  Teddy Lee and
                  Jan Leike and
                  John Schulman and
                  Ilya Sutskever and
                  Karl Cobbe},
  title        = {Let's Verify Step by Step},
  booktitle    = {The Twelfth International Conference on Learning Representations,
                  {ICLR} 2024, Vienna, Austria, May 7-11, 2024},
  publisher    = {OpenReview.net},
  year         = {2024},
  url          = {https://openreview.net/forum?id=v8L0pN6EOi},
  bibsource    = {dblp computer science bibliography, https://dblp.org}
}

@misc{DBLP:journals/corr/abs-2406-03816,
  author       = {Dan Zhang and
                  Sining Zhoubian and
                  Yisong Yue and
                  Yuxiao Dong and
                  Jie Tang},
  title        = {ReST-MCTS*: {LLM} Self-Training via Process Reward Guided Tree Search},
  journal      = {CoRR},
  volume       = {abs/2406.03816},
  year         = {2024},
  url          = {https://doi.org/10.48550/arXiv.2406.03816},
  doi          = {10.48550/ARXIV.2406.03816},
  eprinttype    = {arXiv},
  eprint       = {2406.03816},
  bibsource    = {dblp computer science bibliography, https://dblp.org}
}

@misc{DBLP:conf/nips/ChenL0024a,
  author       = {Guoxin Chen and
                  Minpeng Liao and
                  Chengxi Li and
                  Kai Fan},
  editor       = {Amir Globersons and
                  Lester Mackey and
                  Danielle Belgrave and
                  Angela Fan and
                  Ulrich Paquet and
                  Jakub M. Tomczak and
                  Cheng Zhang},
  title        = {AlphaMath Almost Zero: Process Supervision without Process},
  booktitle    = {Advances in Neural Information Processing Systems 38: Annual Conference
                  on Neural Information Processing Systems 2024, NeurIPS 2024, Vancouver,
                  BC, Canada, December 10 - 15, 2024},
  year         = {2024},
  url          = {http://papers.nips.cc/paper\_files/paper/2024/hash/30dfe47a3ccbee68cffa0c19ccb1bc00-Abstract-Conference.html},
  bibsource    = {dblp computer science bibliography, https://dblp.org}
}

@misc{DBLP:journals/corr/abs-2410-11287,
  author       = {Wendi Li and
                  Yixuan Li},
  title        = {Process Reward Model with Q-Value Rankings},
  journal      = {CoRR},
  volume       = {abs/2410.11287},
  year         = {2024},
  url          = {https://doi.org/10.48550/arXiv.2410.11287},
  doi          = {10.48550/ARXIV.2410.11287},
  eprinttype    = {arXiv},
  eprint       = {2410.11287},
  bibsource    = {dblp computer science bibliography, https://dblp.org}
}

@misc{DBLP:journals/corr/abs-2410-08146,
  author       = {Amrith Setlur and
                  Chirag Nagpal and
                  Adam Fisch and
                  Xinyang Geng and
                  Jacob Eisenstein and
                  Rishabh Agarwal and
                  Alekh Agarwal and
                  Jonathan Berant and
                  Aviral Kumar},
  title        = {Rewarding Progress: Scaling Automated Process Verifiers for {LLM}
                  Reasoning},
  journal      = {CoRR},
  volume       = {abs/2410.08146},
  year         = {2024},
  url          = {https://doi.org/10.48550/arXiv.2410.08146},
  doi          = {10.48550/ARXIV.2410.08146},
  eprinttype    = {arXiv},
  eprint       = {2410.08146},
  bibsource    = {dblp computer science bibliography, https://dblp.org}
}

@misc{DBLP:conf/iclr/Chevalier-Boisvert19,
  author       = {Maxime Chevalier{-}Boisvert and
                  Dzmitry Bahdanau and
                  Salem Lahlou and
                  Lucas Willems and
                  Chitwan Saharia and
                  Thien Huu Nguyen and
                  Yoshua Bengio},
  title        = {BabyAI: {A} Platform to Study the Sample Efficiency of Grounded Language
                  Learning},
  booktitle    = {7th International Conference on Learning Representations, {ICLR} 2019,
                  New Orleans, LA, USA, May 6-9, 2019},
  publisher    = {OpenReview.net},
  year         = {2019},
  url          = {https://openreview.net/forum?id=rJeXCo0cYX},
  bibsource    = {dblp computer science bibliography, https://dblp.org}
}

@misc{DBLP:journals/corr/abs-2406-06592,
  author       = {Liangchen Luo and
                  Yinxiao Liu and
                  Rosanne Liu and
                  Samrat Phatale and
                  Harsh Lara and
                  Yunxuan Li and
                  Lei Shu and
                  Yun Zhu and
                  Lei Meng and
                  Jiao Sun and
                  Abhinav Rastogi},
  title        = {Improve Mathematical Reasoning in Language Models by Automated Process
                  Supervision},
  journal      = {CoRR},
  volume       = {abs/2406.06592},
  year         = {2024},
  url          = {https://doi.org/10.48550/arXiv.2406.06592},
  doi          = {10.48550/ARXIV.2406.06592},
  eprinttype    = {arXiv},
  eprint       = {2406.06592},
  bibsource    = {dblp computer science bibliography, https://dblp.org}
}

@misc{DBLP:conf/iclr/0036YZXLL0DMYZ024,
  author       = {Xiao Liu and
                  Hao Yu and
                  Hanchen Zhang and
                  Yifan Xu and
                  Xuanyu Lei and
                  Hanyu Lai and
                  Yu Gu and
                  Hangliang Ding and
                  Kaiwen Men and
                  Kejuan Yang and
                  Shudan Zhang and
                  Xiang Deng and
                  Aohan Zeng and
                  Zhengxiao Du and
                  Chenhui Zhang and
                  Sheng Shen and
                  Tianjun Zhang and
                  Yu Su and
                  Huan Sun and
                  Minlie Huang and
                  Yuxiao Dong and
                  Jie Tang},
  title        = {AgentBench: Evaluating LLMs as Agents},
  booktitle    = {The Twelfth International Conference on Learning Representations,
                  {ICLR} 2024, Vienna, Austria, May 7-11, 2024},
  publisher    = {OpenReview.net},
  year         = {2024},
  url          = {https://openreview.net/forum?id=zAdUB0aCTQ},
  bibsource    = {dblp computer science bibliography, https://dblp.org}
}

@misc{sutton2018reinforcement,
  title={Reinforcement learning: An introduction},
  author={Sutton, Richard S and Barto, Andrew G},
  year={2018},
  publisher={MIT press}
}

@misc{sutton1999policy,
  title={Policy gradient methods for reinforcement learning with function approximation},
  author={Sutton, Richard S and McAllester, David and Singh, Satinder and Mansour, Yishay},
  journal={Advances in neural information processing systems},
  volume={12},
  year={1999}
}

@misc{sutton1988learning,
  title={Learning to predict by the methods of temporal differences},
  author={Sutton, Richard S},
  journal={Machine learning},
  volume={3},
  pages={9--44},
  year={1988},
  publisher={Springer}
}

@misc{DBLP:journals/corr/abs-2310-05915,
  author       = {Baian Chen and
                  Chang Shu and
                  Ehsan Shareghi and
                  Nigel Collier and
                  Karthik Narasimhan and
                  Shunyu Yao},
  title        = {FireAct: Toward Language Agent Fine-tuning},
  journal      = {CoRR},
  volume       = {abs/2310.05915},
  year         = {2023},
  url          = {https://doi.org/10.48550/arXiv.2310.05915},
  doi          = {10.48550/ARXIV.2310.05915},
  eprinttype    = {arXiv},
  eprint       = {2310.05915},
  bibsource    = {dblp computer science bibliography, https://dblp.org}
}

@misc{ARMAP2024,
  author       = {Zhenfang Chen and Delin Chen and Rui Sun and Wenjun Liu and Chuang Gan},
  title        = {Inference-Time Scaling of Autonomous Agents from Automatic Reward Modeling And Planning},
  year         = {2025},
  month        = {1},
  url          = {https://github.com/heaplax/ARMAP},
}

@misc{DBLP:journals/corr/abs-2306-02224,
  author       = {Hui Yang and
                  Sifu Yue and
                  Yunzhong He},
  title        = {Auto-GPT for Online Decision Making: Benchmarks and Additional Opinions},
  journal      = {CoRR},
  volume       = {abs/2306.02224},
  year         = {2023},
  url          = {https://doi.org/10.48550/arXiv.2306.02224},
  doi          = {10.48550/ARXIV.2306.02224},
  eprinttype    = {arXiv},
  eprint       = {2306.02224},
  bibsource    = {dblp computer science bibliography, https://dblp.org}
}

@misc{DBLP:conf/naacl/YuGW24,
  author       = {Fei Yu and
                  Anningzhe Gao and
                  Benyou Wang},
  editor       = {Kevin Duh and
                  Helena G{\'{o}}mez{-}Adorno and
                  Steven Bethard},
  title        = {OVM, Outcome-supervised Value Models for Planning in Mathematical
                  Reasoning},
  booktitle    = {Findings of the Association for Computational Linguistics: {NAACL}
                  2024, Mexico City, Mexico, June 16-21, 2024},
  pages        = {858--875},
  publisher    = {Association for Computational Linguistics},
  year         = {2024},
  url          = {https://doi.org/10.18653/v1/2024.findings-naacl.55},
  doi          = {10.18653/V1/2024.FINDINGS-NAACL.55},
  bibsource    = {dblp computer science bibliography, https://dblp.org}
}

@misc{DBLP:journals/corr/abs-2407-01476,
  author       = {Jing Yu Koh and
                  Stephen McAleer and
                  Daniel Fried and
                  Ruslan Salakhutdinov},
  title        = {Tree Search for Language Model Agents},
  journal      = {CoRR},
  volume       = {abs/2407.01476},
  year         = {2024},
  url          = {https://doi.org/10.48550/arXiv.2407.01476},
  doi          = {10.48550/ARXIV.2407.01476},
  eprinttype    = {arXiv},
  eprint       = {2407.01476},
  bibsource    = {dblp computer science bibliography, https://dblp.org}
}

@misc{DBLP:conf/emnlp/HaoGMHWWH23,
  author       = {Shibo Hao and
                  Yi Gu and
                  Haodi Ma and
                  Joshua Jiahua Hong and
                  Zhen Wang and
                  Daisy Zhe Wang and
                  Zhiting Hu},
  editor       = {Houda Bouamor and
                  Juan Pino and
                  Kalika Bali},
  title        = {Reasoning with Language Model is Planning with World Model},
  booktitle    = {Proceedings of the 2023 Conference on Empirical Methods in Natural
                  Language Processing, {EMNLP} 2023, Singapore, December 6-10, 2023},
  pages        = {8154--8173},
  publisher    = {Association for Computational Linguistics},
  year         = {2023},
  url          = {https://doi.org/10.18653/v1/2023.emnlp-main.507},
  doi          = {10.18653/V1/2023.EMNLP-MAIN.507},
  bibsource    = {dblp computer science bibliography, https://dblp.org}
}

@misc{barron1989bellman,
  title={The Bellman equation for minimizing the maximum cost.},
  author={Barron, EN and Ishii, H},
  journal={NONLINEAR ANAL. THEORY METHODS APPLIC.},
  volume={13},
  number={9},
  pages={1067--1090},
  year={1989}
}

@misc{DBLP:journals/corr/abs-2409-09345,
  author       = {Yuanzhao Zhai and
                  Tingkai Yang and
                  Kele Xu and
                  Dawei Feng and
                  Cheng Yang and
                  Bo Ding and
                  Huaimin Wang},
  title        = {Enhancing Decision-Making for {LLM} Agents via Step-Level Q-Value
                  Models},
  journal      = {CoRR},
  volume       = {abs/2409.09345},
  year         = {2024},
  url          = {https://doi.org/10.48550/arXiv.2409.09345},
  doi          = {10.48550/ARXIV.2409.09345},
  eprinttype    = {arXiv},
  eprint       = {2409.09345},
  bibsource    = {dblp computer science bibliography, https://dblp.org}
}

@misc{DBLP:conf/nips/RafailovSMMEF23,
  author       = {Rafael Rafailov and
                  Archit Sharma and
                  Eric Mitchell and
                  Christopher D. Manning and
                  Stefano Ermon and
                  Chelsea Finn},
  editor       = {Alice Oh and
                  Tristan Naumann and
                  Amir Globerson and
                  Kate Saenko and
                  Moritz Hardt and
                  Sergey Levine},
  title        = {Direct Preference Optimization: Your Language Model is Secretly a
                  Reward Model},
  booktitle    = {Advances in Neural Information Processing Systems 36: Annual Conference
                  on Neural Information Processing Systems 2023, NeurIPS 2023, New Orleans,
                  LA, USA, December 10 - 16, 2023},
  year         = {2023},
  url          = {http://papers.nips.cc/paper\_files/paper/2023/hash/a85b405ed65c6477a4fe8302b5e06ce7-Abstract-Conference.html},
  bibsource    = {dblp computer science bibliography, https://dblp.org}
}

@misc{DBLP:conf/naacl/PrasadKHCSBK24,
  author       = {Archiki Prasad and
                  Alexander Koller and
                  Mareike Hartmann and
                  Peter Clark and
                  Ashish Sabharwal and
                  Mohit Bansal and
                  Tushar Khot},
  editor       = {Kevin Duh and
                  Helena G{\'{o}}mez{-}Adorno and
                  Steven Bethard},
  title        = {ADaPT: As-Needed Decomposition and Planning with Language Models},
  booktitle    = {Findings of the Association for Computational Linguistics: {NAACL}
                  2024, Mexico City, Mexico, June 16-21, 2024},
  pages        = {4226--4252},
  publisher    = {Association for Computational Linguistics},
  year         = {2024},
  url          = {https://doi.org/10.18653/v1/2024.findings-naacl.264},
  doi          = {10.18653/V1/2024.FINDINGS-NAACL.264},
  bibsource    = {dblp computer science bibliography, https://dblp.org}
}

@misc{DBLP:conf/iclr/YaoZYDSN023,
  author       = {Shunyu Yao and
                  Jeffrey Zhao and
                  Dian Yu and
                  Nan Du and
                  Izhak Shafran and
                  Karthik R. Narasimhan and
                  Yuan Cao},
  title        = {ReAct: Synergizing Reasoning and Acting in Language Models},
  booktitle    = {The Eleventh International Conference on Learning Representations,
                  {ICLR} 2023, Kigali, Rwanda, May 1-5, 2023},
  publisher    = {OpenReview.net},
  year         = {2023},
  url          = {https://openreview.net/forum?id=WE\_vluYUL-X},
  bibsource    = {dblp computer science bibliography, https://dblp.org}
}

@misc{DBLP:journals/corr/abs-2408-03314,
  author       = {Charlie Snell and
                  Jaehoon Lee and
                  Kelvin Xu and
                  Aviral Kumar},
  title        = {Scaling {LLM} Test-Time Compute Optimally can be More Effective than
                  Scaling Model Parameters},
  journal      = {CoRR},
  volume       = {abs/2408.03314},
  year         = {2024},
  url          = {https://doi.org/10.48550/arXiv.2408.03314},
  doi          = {10.48550/ARXIV.2408.03314},
  eprinttype    = {arXiv},
  eprint       = {2408.03314},
  bibsource    = {dblp computer science bibliography, https://dblp.org}
}

@misc{DBLP:journals/corr/abs-2408-16737,
  author       = {Hritik Bansal and
                  Arian Hosseini and
                  Rishabh Agarwal and
                  Vinh Q. Tran and
                  Mehran Kazemi},
  title        = {Smaller, Weaker, Yet Better: Training {LLM} Reasoners via Compute-Optimal
                  Sampling},
  journal      = {CoRR},
  volume       = {abs/2408.16737},
  year         = {2024},
  url          = {https://doi.org/10.48550/arXiv.2408.16737},
  doi          = {10.48550/ARXIV.2408.16737},
  eprinttype    = {arXiv},
  eprint       = {2408.16737},
  bibsource    = {dblp computer science bibliography, https://dblp.org}
}

@misc{wang2025examining,
    title={Examining False Positives under Inference Scaling for Mathematical Reasoning},
    author={Yu Wang and Nan Yang and Liang Wang and Furu Wei},
    year={2025},
    eprint={2502.06217},
    archivePrefix={arXiv},
    primaryClass={cs.CL}
}

@misc{DBLP:journals/corr/abs-2401-06080,
  author       = {Binghai Wang and
                  Rui Zheng and
                  Lu Chen and
                  Yan Liu and
                  Shihan Dou and
                  Caishuang Huang and
                  Wei Shen and
                  Senjie Jin and
                  Enyu Zhou and
                  Chenyu Shi and
                  Songyang Gao and
                  Nuo Xu and
                  Yuhao Zhou and
                  Xiaoran Fan and
                  Zhiheng Xi and
                  Jun Zhao and
                  Xiao Wang and
                  Tao Ji and
                  Hang Yan and
                  Lixing Shen and
                  Zhan Chen and
                  Tao Gui and
                  Qi Zhang and
                  Xipeng Qiu and
                  Xuanjing Huang and
                  Zuxuan Wu and
                  Yu{-}Gang Jiang},
  title        = {Secrets of {RLHF} in Large Language Models Part {II:} Reward Modeling},
  journal      = {CoRR},
  volume       = {abs/2401.06080},
  year         = {2024},
  url          = {https://doi.org/10.48550/arXiv.2401.06080},
  doi          = {10.48550/ARXIV.2401.06080},
  eprinttype    = {arXiv},
  eprint       = {2401.06080},
  bibsource    = {dblp computer science bibliography, https://dblp.org}
}

@misc{DBLP:journals/corr/abs-2309-07864,
  author       = {Zhiheng Xi and
                  Wenxiang Chen and
                  Xin Guo and
                  Wei He and
                  Yiwen Ding and
                  Boyang Hong and
                  Ming Zhang and
                  Junzhe Wang and
                  Senjie Jin and
                  Enyu Zhou and
                  Rui Zheng and
                  Xiaoran Fan and
                  Xiao Wang and
                  Limao Xiong and
                  Yuhao Zhou and
                  Weiran Wang and
                  Changhao Jiang and
                  Yicheng Zou and
                  Xiangyang Liu and
                  Zhangyue Yin and
                  Shihan Dou and
                  Rongxiang Weng and
                  Wensen Cheng and
                  Qi Zhang and
                  Wenjuan Qin and
                  Yongyan Zheng and
                  Xipeng Qiu and
                  Xuanjing Huang and
                  Tao Gui},
  title        = {The Rise and Potential of Large Language Model Based Agents: {A} Survey},
  journal      = {CoRR},
  volume       = {abs/2309.07864},
  year         = {2023},
  url          = {https://doi.org/10.48550/arXiv.2309.07864},
  doi          = {10.48550/ARXIV.2309.07864},
  eprinttype    = {arXiv},
  eprint       = {2309.07864},
  bibsource    = {dblp computer science bibliography, https://dblp.org}
}

@misc{QAgent2025,
  author       = {Zongyu Lin and Yao Tang and Da Yin and Stuart X. Yao and Ziniu Hu and Yizhou Sun and Kai-Wei Chang},
  title        = {Q* Agent: Optimizing Language Agents with Q-Guided Exploration},
  year         = {2024},
  month        = {12},
  url          = {https://openreview.net/forum?id=rxUz2DaulF},
}

@misc{
zelikman2022star,
title={{ST}aR: Bootstrapping Reasoning With Reasoning},
author={Eric Zelikman and Yuhuai Wu and Jesse Mu and Noah Goodman},
booktitle={Advances in Neural Information Processing Systems},
editor={Alice H. Oh and Alekh Agarwal and Danielle Belgrave and Kyunghyun Cho},
year={2022},
url={https://openreview.net/forum?id=_3ELRdg2sgI}
}

@misc{yuan2023scalingrelationshiplearningmathematical,
      title={Scaling Relationship on Learning Mathematical Reasoning with Large Language Models}, 
      author={Zheng Yuan and Hongyi Yuan and Chengpeng Li and Guanting Dong and Keming Lu and Chuanqi Tan and Chang Zhou and Jingren Zhou},
      year={2023},
      eprint={2308.01825},
      archivePrefix={arXiv},
      primaryClass={cs.CL},
      url={https://arxiv.org/abs/2308.01825}, 
}

@misc{trung-etal-2024-reft,
    title = "{R}e{FT}: Reasoning with Reinforced Fine-Tuning",
    author = "Trung, Luong  and
      Zhang, Xinbo  and
      Jie, Zhanming  and
      Sun, Peng  and
      Jin, Xiaoran  and
      Li, Hang",
    editor = "Ku, Lun-Wei  and
      Martins, Andre  and
      Srikumar, Vivek",
    booktitle = "Proceedings of the 62nd Annual Meeting of the Association for Computational Linguistics (Volume 1: Long Papers)",
    month = aug,
    year = "2024",
    address = "Bangkok, Thailand",
    publisher = "Association for Computational Linguistics",
    url = "https://aclanthology.org/2024.acl-long.410/",
    doi = "10.18653/v1/2024.acl-long.410",
    pages = "7601--7614"
}

@misc{chen2025betterprocesssupervisionbidirectional,
      title={Better Process Supervision with Bi-directional Rewarding Signals}, 
      author={Wenxiang Chen and Wei He and Zhiheng Xi and Honglin Guo and Boyang Hong and Jiazheng Zhang and Rui Zheng and Nijun Li and Tao Gui and Yun Li and Qi Zhang and Xuanjing Huang},
      year={2025},
      eprint={2503.04618},
      archivePrefix={arXiv},
      primaryClass={cs.CL},
      url={https://arxiv.org/abs/2503.04618}, 
}

@misc{xi2024traininglargelanguagemodels,
      title={Training Large Language Models for Reasoning through Reverse Curriculum Reinforcement Learning}, 
      author={Zhiheng Xi and Wenxiang Chen and Boyang Hong and Senjie Jin and Rui Zheng and Wei He and Yiwen Ding and Shichun Liu and Xin Guo and Junzhe Wang and Honglin Guo and Wei Shen and Xiaoran Fan and Yuhao Zhou and Shihan Dou and Xiao Wang and Xinbo Zhang and Peng Sun and Tao Gui and Qi Zhang and Xuanjing Huang},
      year={2024},
      eprint={2402.05808},
      archivePrefix={arXiv},
      primaryClass={cs.AI},
      url={https://arxiv.org/abs/2402.05808}, 
}

@misc{xiong2024watchstepllmagent,
      title={Watch Every Step! LLM Agent Learning via Iterative Step-Level Process Refinement}, 
      author={Weimin Xiong and Yifan Song and Xiutian Zhao and Wenhao Wu and Xun Wang and Ke Wang and Cheng Li and Wei Peng and Sujian Li},
      year={2024},
      eprint={2406.11176},
      archivePrefix={arXiv},
      primaryClass={cs.CL},
      url={https://arxiv.org/abs/2406.11176}, 
}

@misc{xia2025agentrmenhancingagentgeneralization,
      title={AgentRM: Enhancing Agent Generalization with Reward Modeling}, 
      author={Yu Xia and Jingru Fan and Weize Chen and Siyu Yan and Xin Cong and Zhong Zhang and Yaxi Lu and Yankai Lin and Zhiyuan Liu and Maosong Sun},
      year={2025},
      eprint={2502.18407},
      archivePrefix={arXiv},
      primaryClass={cs.CL},
      url={https://arxiv.org/abs/2502.18407}, 
}

@misc{miao2025boostingvirtualagentlearning,
      title={Boosting Virtual Agent Learning and Reasoning: A Step-wise, Multi-dimensional, and Generalist Reward Model with Benchmark}, 
      author={Bingchen Miao and Yang Wu and Minghe Gao and Qifan Yu and Wendong Bu and Wenqiao Zhang and Yunfei Li and Siliang Tang and Tat-Seng Chua and Juncheng Li},
      year={2025},
      eprint={2503.18665},
      archivePrefix={arXiv},
      primaryClass={cs.CV},
      url={https://arxiv.org/abs/2503.18665}, 
}

@misc{wang2025visualprmeffectiveprocessreward,
      title={VisualPRM: An Effective Process Reward Model for Multimodal Reasoning}, 
      author={Weiyun Wang and Zhangwei Gao and Lianjie Chen and Zhe Chen and Jinguo Zhu and Xiangyu Zhao and Yangzhou Liu and Yue Cao and Shenglong Ye and Xizhou Zhu and Lewei Lu and Haodong Duan and Yu Qiao and Jifeng Dai and Wenhai Wang},
      year={2025},
      eprint={2503.10291},
      archivePrefix={arXiv},
      primaryClass={cs.CV},
      url={https://arxiv.org/abs/2503.10291}, 
}

@misc{aksitov2023restmeetsreactselfimprovement,
      title={ReST meets ReAct: Self-Improvement for Multi-Step Reasoning LLM Agent}, 
      author={Renat Aksitov and Sobhan Miryoosefi and Zonglin Li and Daliang Li and Sheila Babayan and Kavya Kopparapu and Zachary Fisher and Ruiqi Guo and Sushant Prakash and Pranesh Srinivasan and Manzil Zaheer and Felix Yu and Sanjiv Kumar},
      year={2023},
      eprint={2312.10003},
      archivePrefix={arXiv},
      primaryClass={cs.CL},
      url={https://arxiv.org/abs/2312.10003}, 
}

@misc{song2024trialerrorexplorationbasedtrajectory,
      title={Trial and Error: Exploration-Based Trajectory Optimization for LLM Agents}, 
      author={Yifan Song and Da Yin and Xiang Yue and Jie Huang and Sujian Li and Bill Yuchen Lin},
      year={2024},
      eprint={2403.02502},
      archivePrefix={arXiv},
      primaryClass={cs.CL},
      url={https://arxiv.org/abs/2403.02502}, 
}

@misc{yang2024reactmeetsactrelanguage,
      title={ReAct Meets ActRe: When Language Agents Enjoy Training Data Autonomy}, 
      author={Zonghan Yang and Peng Li and Ming Yan and Ji Zhang and Fei Huang and Yang Liu},
      year={2024},
      eprint={2403.14589},
      archivePrefix={arXiv},
      primaryClass={cs.AI},
      url={https://arxiv.org/abs/2403.14589}, 
}

@misc{tao2024surveyselfevolutionlargelanguage,
      title={A Survey on Self-Evolution of Large Language Models}, 
      author={Zhengwei Tao and Ting-En Lin and Xiancai Chen and Hangyu Li and Yuchuan Wu and Yongbin Li and Zhi Jin and Fei Huang and Dacheng Tao and Jingren Zhou},
      year={2024},
      eprint={2404.14387},
      archivePrefix={arXiv},
      primaryClass={cs.CL},
      url={https://arxiv.org/abs/2404.14387}, 
}

@misc{deng2023mind2webgeneralistagentweb,
      title={Mind2Web: Towards a Generalist Agent for the Web}, 
      author={Xiang Deng and Yu Gu and Boyuan Zheng and Shijie Chen and Samuel Stevens and Boshi Wang and Huan Sun and Yu Su},
      year={2023},
      eprint={2306.06070},
      archivePrefix={arXiv},
      primaryClass={cs.CL},
      url={https://arxiv.org/abs/2306.06070}, 
}

@misc{xu2021groundingopendomaininstructionsautomate,
      title={Grounding Open-Domain Instructions to Automate Web Support Tasks}, 
      author={Nancy Xu and Sam Masling and Michael Du and Giovanni Campagna and Larry Heck and James Landay and Monica S Lam},
      year={2021},
      eprint={2103.16057},
      archivePrefix={arXiv},
      primaryClass={cs.CL},
      url={https://arxiv.org/abs/2103.16057}, 
}

@misc{koh2024visualwebarenaevaluatingmultimodalagents,
      title={VisualWebArena: Evaluating Multimodal Agents on Realistic Visual Web Tasks}, 
      author={Jing Yu Koh and Robert Lo and Lawrence Jang and Vikram Duvvur and Ming Chong Lim and Po-Yu Huang and Graham Neubig and Shuyan Zhou and Ruslan Salakhutdinov and Daniel Fried},
      year={2024},
      eprint={2401.13649},
      archivePrefix={arXiv},
      primaryClass={cs.LG},
      url={https://arxiv.org/abs/2401.13649}, 
}

@misc{zhang2025shopr1rewardingllmssimulate,
      title={Shop-R1: Rewarding LLMs to Simulate Human Behavior in Online Shopping via Reinforcement Learning}, 
      author={Yimeng Zhang and Tian Wang and Jiri Gesi and Ziyi Wang and Yuxuan Lu and Jiacheng Lin and Sinong Zhan and Vianne Gao and Ruochen Jiao and Junze Liu and Kun Qian and Yuxin Tang and Ran Xue and Houyu Zhang and Qingjun Cui and Yufan Guo and Dakuo Wang},
      year={2025},
      eprint={2507.17842},
      archivePrefix={arXiv},
      primaryClass={cs.CL},
      url={https://arxiv.org/abs/2507.17842}, 
}

@misc{wang2025digitalplayerevaluatinglarge,
      title={Digital Player: Evaluating Large Language Models based Human-like Agent in Games}, 
      author={Jiawei Wang and Kai Wang and Shaojie Lin and Runze Wu and Bihan Xu and Lingeng Jiang and Shiwei Zhao and Renyu Zhu and Haoyu Liu and Zhipeng Hu and Zhong Fan and Le Li and Tangjie Lyu and Changjie Fan},
      year={2025},
      eprint={2502.20807},
      archivePrefix={arXiv},
      primaryClass={cs.LG},
      url={https://arxiv.org/abs/2502.20807}, 
}

@misc{park2025orakfoundationalbenchmarktraining,
      title={Orak: A Foundational Benchmark for Training and Evaluating LLM Agents on Diverse Video Games}, 
      author={Dongmin Park and Minkyu Kim and Beongjun Choi and Junhyuck Kim and Keon Lee and Jonghyun Lee and Inkyu Park and Byeong-Uk Lee and Jaeyoung Hwang and Jaewoo Ahn and Ameya S. Mahabaleshwarkar and Bilal Kartal and Pritam Biswas and Yoshi Suhara and Kangwook Lee and Jaewoong Cho},
      year={2025},
      eprint={2506.03610},
      archivePrefix={arXiv},
      primaryClass={cs.AI},
      url={https://arxiv.org/abs/2506.03610}, 
}

@misc{hu2025gamearenaevaluatingllmreasoning,
      title={GameArena: Evaluating LLM Reasoning through Live Computer Games}, 
      author={Lanxiang Hu and Qiyu Li and Anze Xie and Nan Jiang and Ion Stoica and Haojian Jin and Hao Zhang},
      year={2025},
      eprint={2412.06394},
      archivePrefix={arXiv},
      primaryClass={cs.AI},
      url={https://arxiv.org/abs/2412.06394}, 
}

@misc{ding2025mitigatingtailnarrowingllm,
      title={Mitigating Tail Narrowing in LLM Self-Improvement via Socratic-Guided Sampling}, 
      author={Yiwen Ding and Zhiheng Xi and Wei He and Zhuoyuan Li and Yitao Zhai and Xiaowei Shi and Xunliang Cai and Tao Gui and Qi Zhang and Xuanjing Huang},
      year={2025},
      eprint={2411.00750},
      archivePrefix={arXiv},
      primaryClass={cs.CL},
      url={https://arxiv.org/abs/2411.00750}, 
}

@misc{keskar2019ctrlconditionaltransformerlanguage,
      title={CTRL: A Conditional Transformer Language Model for Controllable Generation}, 
      author={Nitish Shirish Keskar and Bryan McCann and Lav R. Varshney and Caiming Xiong and Richard Socher},
      year={2019},
      eprint={1909.05858},
      archivePrefix={arXiv},
      primaryClass={cs.CL},
      url={https://arxiv.org/abs/1909.05858}, 
}

@misc{dathathri2020plugplaylanguagemodels,
      title={Plug and Play Language Models: A Simple Approach to Controlled Text Generation}, 
      author={Sumanth Dathathri and Andrea Madotto and Janice Lan and Jane Hung and Eric Frank and Piero Molino and Jason Yosinski and Rosanne Liu},
      year={2020},
      eprint={1912.02164},
      archivePrefix={arXiv},
      primaryClass={cs.CL},
      url={https://arxiv.org/abs/1912.02164}, 
}

@misc{10.1145/3617680,
      author = {Zhang, Hanqing and Song, Haolin and Li, Shaoyu and Zhou, Ming and Song, Dawei},
      title = {A Survey of Controllable Text Generation Using Transformer-based Pre-trained Language Models},
      year = {2023},
      issue_date = {March 2024},
      publisher = {Association for Computing Machinery},
      address = {New York, NY, USA},
      volume = {56},
      number = {3},
      issn = {0360-0300},
      url = {https://doi.org/10.1145/3617680},
      doi = {10.1145/3617680},
      journal = {ACM Comput. Surv.},
      month = oct,
      miscno = {64},
      numpages = {37}
}

@misc{10.1145/3649449,
      author = {Li, Junyi and Tang, Tianyi and Zhao, Wayne Xin and Nie, Jian-Yun and Wen, Ji-Rong},
      title = {Pre-Trained Language Models for Text Generation: A Survey},
      year = {2024},
      issue_date = {September 2024},
      publisher = {Association for Computing Machinery},
      address = {New York, NY, USA},
      volume = {56},
      number = {9},
      issn = {0360-0300},
      url = {https://doi.org/10.1145/3649449},
      doi = {10.1145/3649449},
      journal = {ACM Comput. Surv.},
      month = apr,
      miscno = {230},
      numpages = {39}
}

@misc{tang2023evaluating,
      title={Evaluating large language models on medical evidence summarization},
      author={Tang, Liyan and Sun, Zhaoyi and Idnay, Betina and Nestor, Jordan G and Soroush, Ali and Elias, Pierre A and Xu, Ziyang and Ding, Ying and Durrett, Greg and Rousseau, Justin F and others},
      journal={NPJ digital medicine},
      volume={6},
      number={1},
      pages={158},
      year={2023},
      publisher={Nature Publishing Group UK London}
}

@misc{10.1162/tacla00632,
      author = {Zhang, Tianyi and Ladhak, Faisal and Durmus, Esin and Liang, Percy and McKeown, Kathleen and Hashimoto, Tatsunori B.},
      title = {Benchmarking Large Language Models for News Summarization},
      journal = {Transactions of the Association for Computational Linguistics},
      volume = {12},
      pages = {39-57},
      year = {2024},
      month = {01},
      issn = {2307-387X},
}

@misc{van2024adapted,
      title={Adapted large language models can outperform medical experts in clinical text summarization},
      author={Van Veen, Dave and Van Uden, Cara and Blankemeier, Louis and Delbrouck, Jean-Benoit and Aali, Asad and Bluethgen, Christian and Pareek, Anuj and Polacin, Malgorzata and Reis, Eduardo Pontes and Seehofnerov{\'a}, Anna and others},
      journal={Nature medicine},
      volume={30},
      number={4},
      pages={1134--1142},
      year={2024},
      publisher={Nature Publishing Group US New York}
}

@misc{zhang2025comprehensivesurveyprocessorientedautomatic,
      title={A Comprehensive Survey on Process-Oriented Automatic Text Summarization with Exploration of LLM-Based Methods}, 
      author={Yang Zhang and Hanlei Jin and Dan Meng and Jun Wang and Jinghua Tan},
      year={2025},
      eprint={2403.02901},
      archivePrefix={arXiv},
      primaryClass={cs.AI},
      url={https://arxiv.org/abs/2403.02901}, 
}

@misc{moslem2023adaptivemachinetranslationlarge,
      title={Adaptive Machine Translation with Large Language Models}, 
      author={Yasmin Moslem and Rejwanul Haque and John D. Kelleher and Andy Way},
      year={2023},
      eprint={2301.13294},
      archivePrefix={arXiv},
      primaryClass={cs.CL},
      url={https://arxiv.org/abs/2301.13294}, 
}

@misc{wang2023documentlevelmachinetranslationlarge,
      title={Document-Level Machine Translation with Large Language Models}, 
      author={Longyue Wang and Chenyang Lyu and Tianbo Ji and Zhirui Zhang and Dian Yu and Shuming Shi and Zhaopeng Tu},
      year={2023},
      eprint={2304.02210},
      archivePrefix={arXiv},
      primaryClass={cs.CL},
      url={https://arxiv.org/abs/2304.02210}, 
}

@misc{pmlr-v202-zhang23m,
      title = 	 {Prompting Large Language Model for Machine Translation: A Case Study},
      author =       {Zhang, Biao and Haddow, Barry and Birch, Alexandra},
      booktitle = 	 {Proceedings of the 40th International Conference on Machine Learning},
      pages = 	 {41092--41110},
      year = 	 {2023},
      editor = 	 {Krause, Andreas and Brunskill, Emma and Cho, Kyunghyun and Engelhardt, Barbara and Sabato, Sivan and Scarlett, Jonathan},
      volume = 	 {202},
      series = 	 {Proceedings of Machine Learning Research},
      month = 	 {23--29 Jul},
      publisher =    {PMLR},
      url = 	 {https://proceedings.mlr.press/v202/zhang23m.html}
}

@misc{xu2024paradigmshiftmachinetranslation,
      title={A Paradigm Shift in Machine Translation: Boosting Translation Performance of Large Language Models}, 
      author={Haoran Xu and Young Jin Kim and Amr Sharaf and Hany Hassan Awadalla},
      year={2024},
      eprint={2309.11674},
      archivePrefix={arXiv},
      primaryClass={cs.CL},
      url={https://arxiv.org/abs/2309.11674}, 
}

@misc{10.1162/tacla00642,
      author = {He, Zhiwei and Liang, Tian and Jiao, Wenxiang and Zhang, Zhuosheng and Yang, Yujiu and Wang, Rui and Tu, Zhaopeng and Shi, Shuming and Wang, Xing},
      title = {Exploring Human-Like Translation Strategy with Large Language
                        Models},
      journal = {Transactions of the Association for Computational Linguistics},
      volume = {12},
      pages = {229-246},
      year = {2024},
      month = {03},
      issn = {2307-387X},
}

@misc{openai2024o1,
  author       = {OpenAI},
  title        = {Introducing OpenAI o1-preview},
  year         = {2024},
  month        = {9},
  url          = {https://openai.com/index/introducing-openai-o1-preview/},
}

@misc{deepseekai2025deepseekr1incentivizingreasoningcapability,
      title={DeepSeek-R1: Incentivizing Reasoning Capability in LLMs via Reinforcement Learning}, 
      author={DeepSeek-AI},
      year={2025},
      eprint={2501.12948},
      archivePrefix={arXiv},
      primaryClass={cs.CL},
      url={https://arxiv.org/abs/2501.12948}, 
}

\clearpage
\newpage

\appendix
\section*{\centering \LARGE{Appendix}}

\section{More Detailed Discussion of Related Work}\label{app: More Detailed Discussion of Related Work}

We list the comparison of our method and other related methods in Table \ref{tab: comparison_PRM_paradigms} of \ref{app: More Detailed Discussion of Related Work}.

In the LLM agent domain, ARMAP constructs outcome reward models (ORMs) through data labeling to re-rank trajectories, providing better BoN performance \citep{ARMAP2024}. Q* AGENT uses the Bellman equation \citep{barron1989bellman} to estimate the Q-value of each step to train process reward models \citep{QAgent2025}. DPO-Q \citep{DBLP:journals/corr/abs-2409-09345} uses a MCTS-based method for building a planning tree and use DPO \citep{DBLP:conf/nips/RafailovSMMEF23} to estimate the value of each step. IPR \citep{xiong2024watchstepllmagent} acquires step-level reward by exploring from every step and calculate the mean reward of the explorations. AgentRM \citep{xia2025agentrmenhancingagentgeneralization} deploys a MCTS-inspired method to collect the trajectories for training and estimate the values of each step with information stored in the search tree. Similar \citep{miao2025boostingvirtualagentlearning} also proposes a MCTS-based dataset collecting method. To better adapt to its real-world tasks, it evaluates the step-level value from 5 dimensions based on the outcome reward as well as information like length of the trajectories. However, they only consider the promise of each step, without accounting for the dependencies and progress between actions. In contrast, our approach uses TD-based estimation with GAE to estimate the value at different steps, capturing the dependencies between actions. 

In the LLM reasoning domain, PQM also considers the relationships between different steps, but unlike us, they use MC-based estimation and introduce a ranking loss to optimize the model \citep{DBLP:journals/corr/abs-2410-11287}. PAV, on the other hand, estimates the reward of the entire trajectory through ORM and incorporates the advantages of individual steps to assist RL and search \citep{DBLP:journals/corr/abs-2410-08146}.

\begin{table}[h]
    \centering
    \caption{Comparison of different process supervision paradigms. ``SG'' means Supervision Granularity, and ``P'' means Progress.}
    \resizebox{0.8\linewidth}{!}{
        \begin{tabular}{llccc}
            \toprule
            \textbf{Method} & \textbf{Labeling} & \textbf{SG} & \textbf{P} & \textbf{Task Type} \\ \midrule
            PRM \citep{DBLP:conf/iclr/LightmanKBEBLLS24} & Human & Process & \texttimes & Reasoning \\ \midrule
            Math-Shepherd \citep{DBLP:conf/acl/WangLSXDLCWS24} & MC-based & Process & \texttimes & Reasoning \\ \midrule
            PAV \citep{DBLP:journals/corr/abs-2410-08146} & MC-based & Process & \checkmark & Reasoning \\ \midrule
            PQM \citep{DBLP:journals/corr/abs-2410-11287} & MC-based & Process & \checkmark & Reasoning \\ \midrule
            ARMAP \citep{ARMAP2024} & MC-based & Outcome & \texttimes & Agent \\ \midrule
            Q* Agent \citep{QAgent2025} & TD-based & Process & \texttimes & Agent \\ \midrule
            DPO-Q \citep{DBLP:journals/corr/abs-2409-09345} & MC-based & Process & \texttimes & Agent \\ \midrule
            IPR \citep{xiong2024watchstepllmagent} & MC-based & Process & \texttimes & Agent \\ \midrule
            AgentRM \citep{xia2025agentrmenhancingagentgeneralization} & MC-based & Process & \texttimes & Agent \\ \midrule
            Similar \citep{miao2025boostingvirtualagentlearning} & MC-based & Process & \texttimes & Agent \\ 
            \midrule
            AgentPRM (Ours) & TD-based & Process & \checkmark & Agent, Reasoning \\
            \bottomrule
        \end{tabular}
    }
    \label{tab: comparison_PRM_paradigms}
\end{table}

\section{Algorithm}\label{app: Algorithm}

We demonstrate the training algorithm of AgentPRM in \ref{algo: Training of AgentPRM}, and the process of beam search in Algorithm \ref{algo: Beam search with AgentPRM}.

\begin{algorithm}[h]
\caption{Training of AgentPRM}
\label{algo: Training of AgentPRM}
  \SetKwData{Left}{left}\SetKwData{This}{this}\SetKwData{Up}{up}
  \SetKwFunction{Union}{Union}\SetKwFunction{FindCompress}{FindCompress}
  \SetKwInOut{Input}{Input}\SetKwInOut{Output}{Output}
   \SetKwProg{myproc}{Procedure}{}{}
   \KwIn{Initialized AgentPRM model $\mathcal{M}_\phi$; Reward function $r$; Sample number per query $N_{\textnormal{TD}}$; Agent task query set $\{s^i_0\}_{i=1}^{N_\textnormal{Task}}$; Actor $\pi_\theta$; Number of training iterations $m$.}
    \myproc{{\textnormal{Trajectories collection}}}{
        $\mathcal{D}_{train} \gets [\  \ ]$
        \Comment{Initialize AgentPRM Train set $\mathcal{D}_{train}$}\\
        \For{$s^i_0$ \textnormal{in} $\{s^i_0\}_{i=1}^{N_\textnormal{Task}}$}{
            \For{$n=1$ \textnormal{to} $N_{\textnormal{TD}}$}{
            $\tau \gets \pi_\theta(s^i_0)$;\\
            Add $\tau$ to $\mathcal{D}_{train}$;
            }
        }
    }
   \myproc{{\textnormal{AgentPRM model training}}}{  
   \For{$n=1$ \textnormal{to} $m$}{
        \For{\textnormal{batch} \textnormal{in}  $\mathcal{D}_{train}$}{
            \For{\textnormal{trajectory} $\tau$ \textnormal{in} \textnormal{batch}}{
            
            $\mathcal{Q} \gets [\  \ ]$; 
            \Comment{AgentPRM model estimated value list $\mathcal{Q}$}\\
            \For{$(s_t,a_t)$ \textnormal{in} $\tau$}{
                
                    $Q_t \gets \mathcal{M}_\phi(s_t,a_t)$ \\
                    Add $Q_t$ to $\mathcal{Q}$; \\   
                }
            $\hat{A} \gets GAE(\mathcal{Q},r(\tau))$; \\
            $\hat{Q} \gets TD(\mathcal{A},\mathcal{Q})$ \\
            $\mathcal{L}_Q=\mathbb{E}\left[ \frac{1}{2}(\mathcal{Q}_t - \hat{Q}_t)^2 \right]$\\
            $\mathcal{L}_A=\mathbb{E}\left[ \frac{1}{2}((\mathcal{Q}_t -\mathcal{Q}_{t-1}) - (\hat{Q}_t-\hat{Q}_{t-1}))^2 \right]$\\
            $\mathcal{M}_\phi \gets \textnormal{Back\_Propagation}(\mathcal{L}_Q + \beta\mathcal{L}_A)$
            }
        }
    }
    }
\end{algorithm}

\begin{algorithm}[t]
\caption{Beam search with PRM.}
\label{algo: Beam search with AgentPRM}
  \SetKwData{Left}{left}\SetKwData{This}{this}\SetKwData{Up}{up}
  \SetKwFunction{Union}{Union}\SetKwFunction{FindCompress}{FindCompress}
  \SetKwInOut{Input}{Input}\SetKwInOut{Output}{Output}
   \SetKwProg{myproc}{Procedure}{}{}
   \KwIn{Trained PRM $\mathcal{M}_\phi$; Policy $\pi_\theta$; Number of actions expanded at each node $M$; Size of beam search $N$; Max steps $T$}
   
   \myproc{{\textnormal{Step-level beam search with PRM}}}{  
   $\mathcal{C} = [\mathbf{s}_0] * M$, $t=0$    \Comment{Initialize candidates}\\
   \While{$t < T$ \textbf{and} non-terminal path in $\mathcal{C}$}{
        $\mathcal{C}_{t+1} \gets [\  \ ]$       \Comment{Initialize priority queue}\\
        \For{$\mathbf{s}_t$ in $\mathcal{C}$}{
            Sample $\left\{\mathbf{a}^{(b)}\right\}_{b=1}^{B_2} \sim \pi_\theta(\mathbf{s}_t)$\\
        \For{$b=1$ to $M$}{          
            $\mathbf{s_{t+1}} = \text{Concat}\left[\mathbf{s}_t, \mathbf{a}^{(b)}\right]$\\
            Add $\left(\mathbf{s}_{t+1}, \mathcal{M}_\phi(\mathbf{s}_{t+1})\right)$ to $\mathcal{C}_{t+1}$ 
            }
        $\mathcal{C} \gets$ Top-$N$ of $\mathcal{C}_{t+1}$
        }
}
    \Return Top-1 of $\mathcal{C}$ \Comment{Return top-1 as the final solution path}
    }
\end{algorithm}

\section{More Implementation Details for RL}\label{app: Implementation Details of Reinforcement Learning}

We train LLM agents with PPO using a batch size of $16$, a learning rate of $1 \times 10^{-6}$, a KL coefficient of $1 \times 10^{-3}$, and a sampling temperature of $1.0$. For both BabyAI and TextCraft, the maximum interaction horizon is set to $20$. In the PRM-based RL, we use the PRM’s predicted score at the final step as the reward for the trajectory.

\section{Qualitative Analysis}\label{app: Qualitative Analysis}
\begin{figure*}[b]
    \centering
    \includegraphics[width=0.9\linewidth]{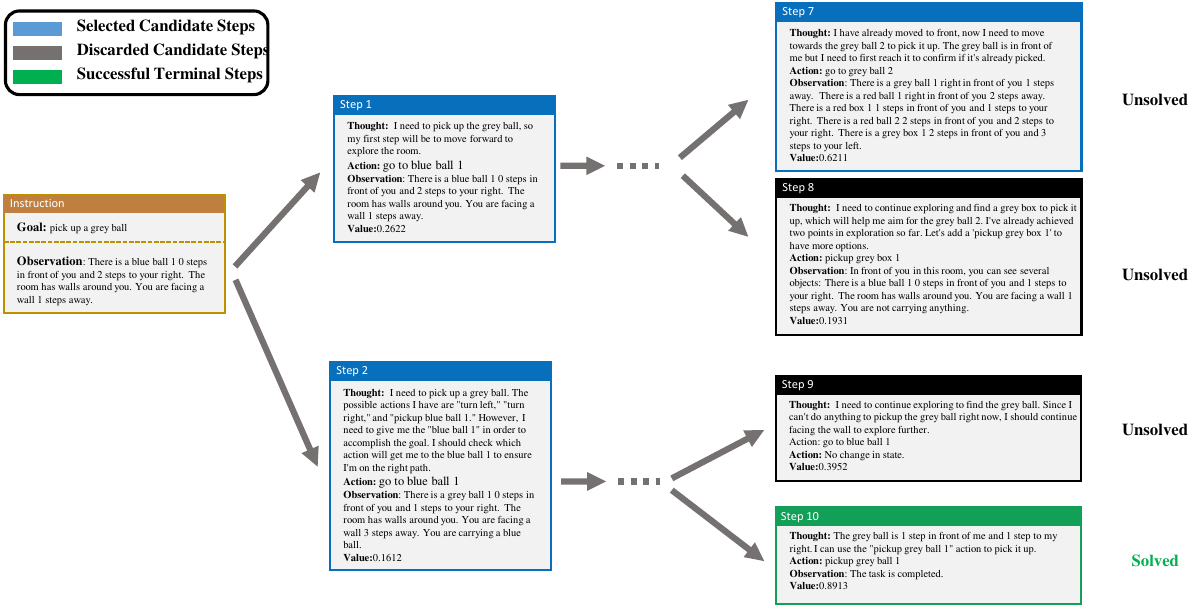}
    \caption{Example of qualitative analysis on beam search. This figure demonstrates a successful solution with beam search guided by AgentPRM. The policy model solves this task in $10$ steps under the guidance of AgentPRM.}
    \label{fig: case_study_beam}
\end{figure*}

We perform a qualitative analysis to show how AgentPRM works. The case shown in Figure \ref{fig: case_study_beam} demonsrates the process of beam search guided by AgentPRM. The LLM agent successfully solves this task under the guidance of AgentPRM. We can also find that AgentPRM effectively distinguishes between good and bad actions (assigning high scores to good actions and low scores to bad ones).

\end{document}